%% file: main.tex
\PassOptionsToPackage{compress}{natbib}
\documentclass{article}
\RequirePackage[table,dvipsnames,svgnames]{xcolor}
\usepackage[preprint]{corl_2025}
\input{config}

\title{\dna \model: Closed-Loop Whole-Body Humanoid Teleoperation for Long-Horizon Tasks}

\author{%
    \normalfont\setlength{\tabcolsep}{3pt}%
    \hspace{-1em}\begin{tabular}{ccc}
        \textbf{Yixuan Li}$^{\,^\ast,1,2}$ & \textbf{Yutang Lin}$^{\,^\ast,3,4,5,6}$ & \textbf{Jieming Cui}$^{\,2,3,4,6}$\\
        \texttt{\footnotesize{}lyx@bit.edu.cn} & \texttt{\footnotesize{}yutang.lin@stu.pku.edu.cn} & \texttt{\footnotesize{}cuijieming@stu.pku.edu.cn}
        \vspace{3pt}\\
        \textbf{Tengyu Liu}$^{\,2,7}$ & \textbf{Wei Liang}\textsuperscript{\Letter}$^{\,,1}$ & \textbf{Yixin Zhu}\textsuperscript{\Letter}$^{\,,3,4,6,8}$\\
        \texttt{\footnotesize{}liutengyu@bigai.ai} & \texttt{\footnotesize{}liangwei@bit.edu.cn} & \texttt{\footnotesize{}yixin.zhu@pku.edu.cn}
        \vspace{3pt}\\
        & \textbf{Siyuan Huang}\textsuperscript{\Letter}$^{\,,2,7}$ &\\
        & \texttt{\footnotesize{}syhuang@bigai.ai} &
    \end{tabular}
    \vspace{6pt}\\
    \footnotesize$^\ast$ Equal contributors\quad{}
    \footnotesize$^1$ School of Computer Science and Technology, Beijing Institute of Technology\\
    \footnotesize$^2$ State Key Laboratory of General Artificial Intelligence, BIGAI\\
    \footnotesize$^3$ School of Psychological and Cognitive Sciences, Peking University\\
    \footnotesize$^4$ Institute for Artificial Intelligence, Peking University\quad{}
    \footnotesize$^5$ Yuanpei College, Peking University\\
    \footnotesize$^6$ Beijing Key Laboratory of Behavior and Mental Health, Peking University\\
    \footnotesize$^7$ Joint Laboratory of Embodied AI and Humanoid Robots, BIGAI \& UniTree Robotics\\
    \footnotesize$^8$ Embodied Intelligence Lab, PKU-Wuhan Institute for Artificial Intelligence\\
    \url{https://humanoid-clone.github.io/}
    \vspace{-6pt}
}

\begin{document}
\maketitle

\begin{center}
    \captionsetup{type=figure}
    \vspace{-18pt}
    \includegraphics[width=\linewidth]{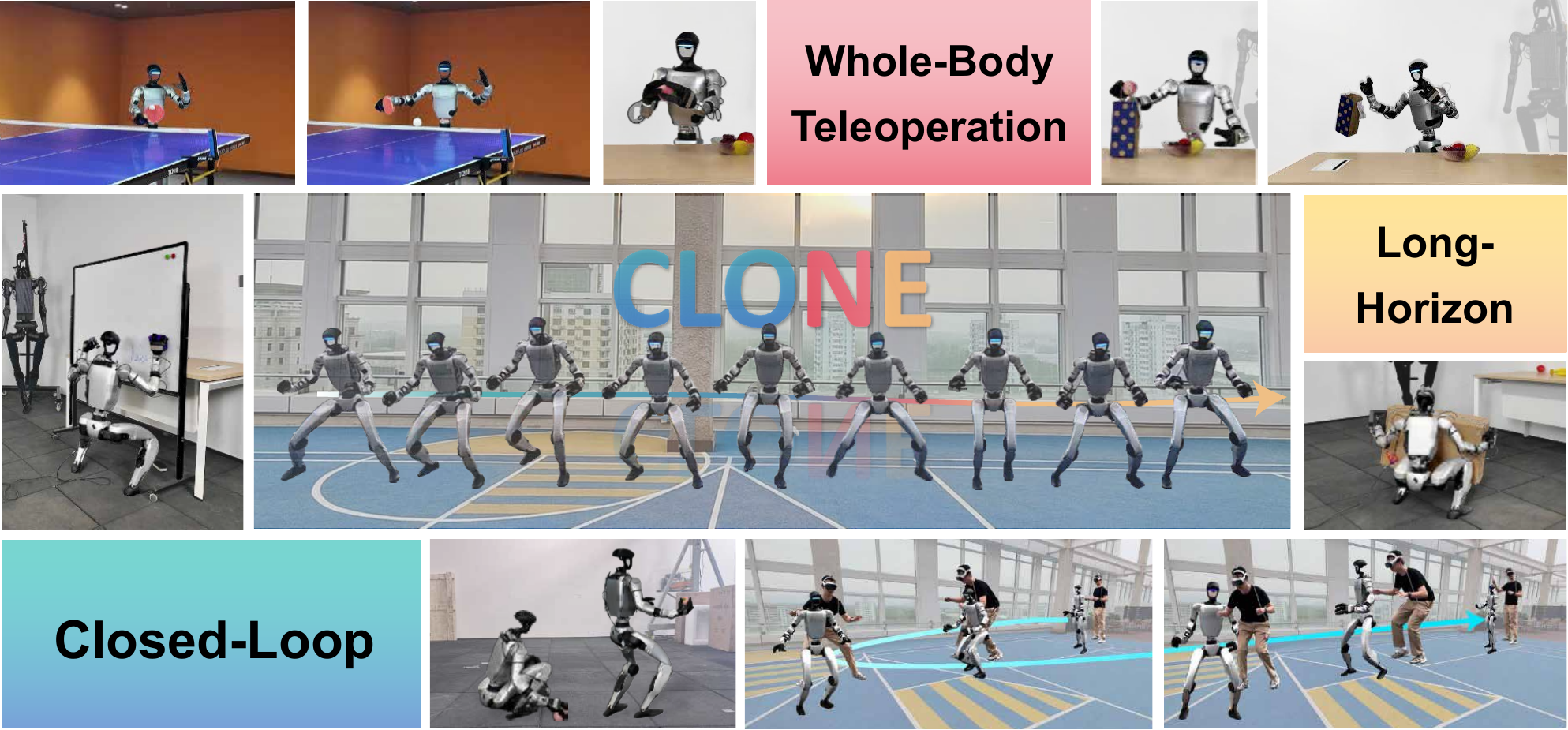}
    \captionof{figure}{\model employs an \acs{moe}-based policy with \textbf{closed-loop} error correction for humanoid teleoperation, enabling precise \textbf{whole-body coordination} and \textbf{long-horizon} task execution.}
    \label{fig:teaser}
    \vspace{-6pt}
\end{center}

\begin{abstract}
Humanoid teleoperation plays a vital role in demonstrating and collecting data for complex humanoid-scene interactions.
However, current teleoperation systems face critical limitations: they decouple upper- and lower-body control to maintain stability, restricting natural coordination, and operate open-loop without real-time position feedback, leading to accumulated drift.
The fundamental challenge is achieving precise, coordinated whole-body teleoperation over extended durations while maintaining accurate global positioning.
Here we show that an \acs{moe}-based teleoperation system, \model, with closed-loop error correction enables unprecedented whole-body teleoperation fidelity, maintaining minimal positional drift over long-range trajectories using only head and hand tracking from an \acs{mr} headset.
Unlike previous methods that either sacrifice coordination for stability or suffer from unbounded drift, \model learns diverse motion skills while preventing tracking error accumulation through real-time feedback, enabling complex coordinated movements such as ``picking up objects from the ground.''
These results establish a new milestone for whole-body humanoid teleoperation for long-horizon humanoid-scene interaction tasks.
\end{abstract}%

\section{Introduction}

\begin{wrapfigure}{r}{0.4\linewidth}
    \centering
    \vspace{-48pt}
    \includegraphics[width=\linewidth]{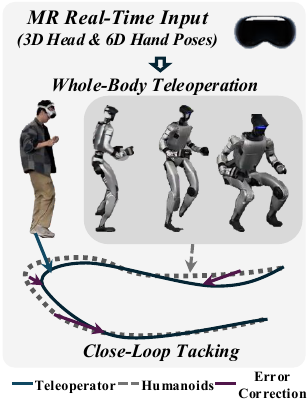}
    \caption{\textbf{Whole-body humanoid teleoperation from minimal input.} Our approach enables intuitive control of a humanoid robot using only head and hand poses from mixed reality input, generating coordinated whole-body motions including natural locomotion. Through closed-loop tracking, the system maintains accurate correspondence between operator and robot over extended operation periods, enabling complex long-horizon tasks that require sustained precision.}
    \label{fig:illustration}
    \vspace{-15pt}
\end{wrapfigure}

The ability to seamlessly coordinate whole-body movements while navigating complex environments represents one of humanity's most remarkable capabilities~\cite{henze2016passivity,wensing2016improved}. From squatting to retrieve objects from the ground to walking across rooms while carrying items, humans effortlessly integrate locomotion and manipulation in ways that remain challenging for robots~\cite{sentis2005synthesis,khatib2003unified,fukuda2017humanoid,hereid2018dynamic,cui2025grove}. Humanoid robots (humanoids henceforth), with their human-like morphology, offer the promise of replicating these capabilities---potentially enabling applications from household assistance to operations in hazardous environments where human-like dexterity and mobility are essential~\cite{tong2024advancements,vecna2006bear,nrl_ash,agibot_a2}.

However, realizing this potential requires solving a fundamental challenge: enabling intuitive and precise teleoperation that maintains coordination across the entire body over extended periods. Long-horizon tasks, such as navigating to distant locations while manipulating objects, demand not only moment-to-moment stability but also sustained accuracy in both movement execution and global positioning. Current teleoperation approaches fall short of these requirements, creating a significant capability gap between human operators and humanoids.

Recent advances in humanoid teleoperation and loco-manipulation~\cite{fu2024humanplus,ze2024generalizable,he2024hover,cui2024anyskill,dafarra2022icub3,ben2025homie,he2024omnih2o,haarnoja2024learning} have made notable progress. Nevertheless, existing methods struggle with precise teleoperation over extended durations and fall short of the whole-body coordination necessary for humanoid-scene interaction. Two fundamental challenges persist in bridging this capability gap.

The first challenge centers on achieving \textbf{coordinated whole-body coordination}. Many systems decouple upper- and lower-body control for stability~\cite{ben2025homie,matsiko2025humanoid}, sacrificing the natural synergies required for fluid motion. While this separation provides safety, it fundamentally limits integrated actions such as reaching while walking or adjusting posture during manipulation. Alternative approaches that rely on motion capture data~\cite{fu2024humanplus,he2024hover,he2024omnih2o,he2024learning,cheng2024expressive,ji2024exbody2,jiang2024scaling,jiang2024autonomous,jiang2025dynamic} often emphasize stability at the cost of expressiveness, yielding conservative motions constrained by training data distributions. Moreover, these methods consistently overlook key factors like hand orientation that are critical for dexterous tasks, further restricting humanoids' potential for sophisticated whole-body movements.

The second challenge involves \textbf{accumulated positional drift} over time due to the absence of real-time feedback about the robot's actual position in the environment. Unlike wheeled robots with straightforward odometry, humanoids exhibit complex foot-ground interactions and non-holonomic dynamics that complicate accurate state estimation. Without closed-loop correction, small pose errors compound with each step, progressively degrading the operator's spatial awareness and control authority, eventually leading to complete task failure. This drift becomes particularly acute during manipulation tasks that require precise positioning relative to environmental objects.

To tackle the above challenges in humanoid long-horizon tasks requiring whole-body coordination and accurate positioning, we present \model. As illustrated in \cref{fig:illustration}, \model is a \textbf{closed-loop} whole-body teleoperation system combining learning-based coordination and real-time feedback correction. Our system employs an \ac{moe} architecture that learns to coordinate diverse motion skills while a LiDAR-based error correction mechanism prevents the accumulation of positional drift. Critically, \model requires only head and hand tracking from a single commercial \ac{mr} headset, making it practical for real-world deployment while achieving unprecedented fidelity in long-horizon tasks.

Our approach integrates three key components: (i) We develop an \ac{moe} framework that enables unified learning of diverse motion skills while maintaining natural upper- and lower-body coordination. (ii) We implement closed-loop error correction using LiDAR odometry~\cite{xu2022fast} and \ac{avp} tracking to provide continuous global pose feedback and prevent drift accumulation. (iii) We curate a comprehensive dataset, \dataset, that augments AMASS~\cite{mahmood2019amass} with additional motion-captured sequences and an online hand orientation generation method, ensuring robust generalization to complex manipulation scenarios involving coordinated whole-body movements.

Our experiments demonstrate that \model enables capabilities previously unattainable with existing systems: whole-body coordination over long trajectories with minimal positional drift, complex coordinated movements like object retrieval from ground level, and robust performance across diverse operator configurations and environmental conditions. Using only minimal input from a commercial \ac{mr} headset, \model achieves improved tracking precision over existing open-loop approaches, opening new possibilities for practical humanoid applications in unstructured environments.

Our contributions are four-fold: (i) the first \ac{moe}-based framework for coordinated whole-body teleoperation that maintains natural movement synergies; (ii) a closed-loop system that solves the fundamental position drift problem in long-horizon tasks through real-time pose correction; (iii) a comprehensive dataset, \dataset, enabling robust learning of dexterous whole-body motions with proper hand orientation coverage; and (iv) extensive validation demonstrating substantial improvements in real-world humanoid-scene interaction capabilities across diverse scenarios.

\section{Related Work}

\paragraph{Whole-Body Humanoid Teleoperation} Humanoid teleoperation enables robots to replicate human movements for complex tasks using motion capture systems~\cite{ji2024exbody2,dafarra2022icub3,darvish2019whole}, haptic devices~\cite{brygo2014humanoid,peternel2013learning,ramos2019dynamic}, or virtual reality interfaces~\cite{he2024hover,he2024omnih2o,chagas2021humanoid,penco2019multimode,tachi2020telesar}. A key challenge is developing whole-body control\footnote{Whole-Body Control (WBC) in robotics literature~\cite{nakamura1987task,khatib2004whole,dietrich2015overview} was traditionally formulated as an optimization problem~\cite{sentis2005synthesis,moro2019whole}, coordinating multiple competing tasks (such as balance and reaching) through hierarchical control objectives. Learning-based approaches~\cite{fu2024humanplus,he2024omnih2o,ji2024exbody2} have recently extended this concept by formulating whole-body control as a reinforcement learning problem. We refer to our approach as a whole-body control policy, as it similarly coordinates all degrees of freedom of the humanoid in a unified manner.} policies that balance robot stability with motion tracking fidelity. Current methods struggle to reproduce the full diversity and fluidity of human motions~\cite{moniruzzaman2022teleoperation}, primarily due to monolithic \acs{mlp}-based architectures that inadequately handle conflicting objectives across different motion types (\eg, walking \vs crouching)~\cite{huang2025moe,zhou2022on,darvish2023teleoperation}. Although mixture-based models have shown promise in other domains~\cite{yang2020multi,xie2022learning,song2024germ,cheng2023multi}, their application to humanoid teleoperation remains underexplored. To address these limitations, we leverage an \ac{moe} framework for adaptive learning and unified representation of diverse motion patterns within a single policy.

\paragraph{Long-Horizon Loco-Manipulation}

Long-horizon task execution~\cite{garrett2021integrated} has been extensively studied for fixed-base arms~\cite{wang2023mimicplay,lin2024hierarchical,shi2023robocook,zhao2024tac}, mobile manipulators~\cite{jiao2021efficieint,jiao2021consolidating,jiao2022planning,zhi2024closedloop,zhi2025learning}, and aerial manipulators~\cite{su2023sequential}, typically in structured settings. In contrast, humanoid teleoperation remains limited to short-horizon motion replication~\cite{he2024omnih2o,cheng2024expressive,ji2024exbody2}, operating open-loop due to difficulties in real-time global state estimation for bipedal systems. Although recent advances in odometry have improved state tracking for legged robots~\cite{wisth2022vilens,ou2024leg,allshire2025visual}, their application to long-horizon humanoid control remains largely unexplored. To bridge this gap, we integrate LiDAR odometry into our teleoperation framework to enable closed-loop error correction and significantly reduce accumulated drift.

\paragraph{Datasets for Training Humanoids}

Large-scale \ac{mocap} datasets~\cite{mahmood2019amass,harvey2020robust} have been instrumental in training humanoid control policies~\cite{he2024hover,he2024omnih2o,cheng2024expressive,ma2025styleloco,geng2025roboverse}. Even after augmenting the datasets with generative models~\cite{ji2024exbody2}, these datasets were still confined primarily to animation and graphics~\cite{wang2022humanise,jiang2024scaling} rather than robotics applications. While they contain semantically distinct actions (\eg, waving, hugging, drinking), they underrepresent the kinematic configurations and dynamic transitions required for robust, generalizable controller training in real-world scenarios. To address these limitations, we introduce \dataset by augmenting AMASS~\cite{mahmood2019amass} through motion editing and collecting additional human \acs{mocap} data, specifically tailored for humanoid controllers. This expansion increases coverage of motions and transitions relevant to humanoid control tasks.

\begin{figure*}[t!]
    \centering
    \includegraphics[width=\linewidth]{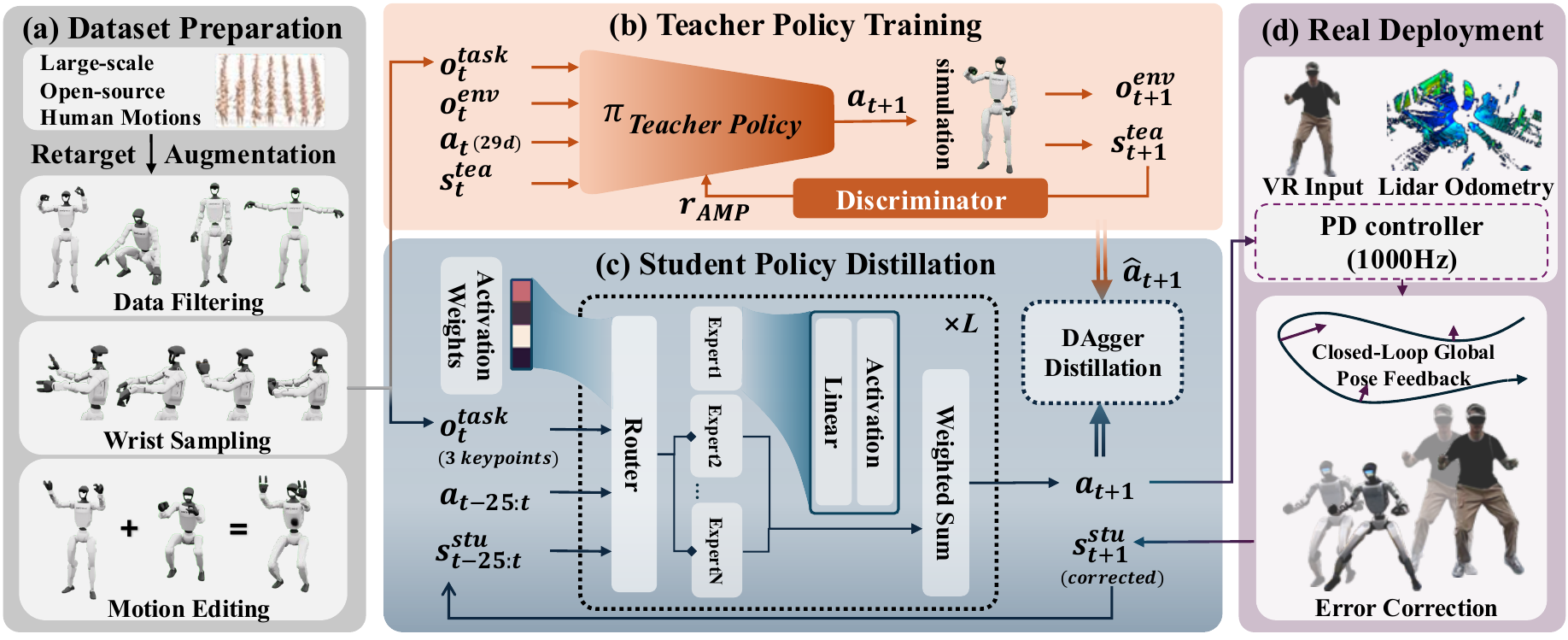}
    \caption{\textbf{The \model framework.} (a) \dataset curates and augments retargeted AMASS~\cite{mahmood2019amass} data through motion editing to introduce diverse humanoid motions and detailed hand movements. (b) A teacher policy is trained using privileged information, including full robot state and environmental context. (c) An \acs{moe} network serves as the student policy, distilled from the teacher to operate with real-world observations only. (d) For real-world deployment, we integrate LiDAR odometry to obtain real-time humanoid states, enabling closed-loop error correction during teleoperation.}
    \label{fig:model}
\end{figure*}

\section{The \model Framework}

Our teleoperation framework captures a minimal set of control signals from the teleoperator, consisting solely of the $6$D poses (position and orientation) of both wrists and the $3$D position of the head, tracked using an \ac{avp} headset. These three points (see also \cref{fig:model}) serve as the complete control interface, providing an intuitive yet powerful means of directing the humanoid's whole-body motion while maintaining a simple setup that requires no additional hardware or complex calibration.

\model addresses two fundamental challenges through complementary components. First, we develop a teacher-student policy learning approach that transforms these sparse control signals into coordinated whole-body movements (\cref{sec:training}). Second, we implement a closed-loop error correction mechanism that maintains positional accuracy during extended operation (\cref{sec:closedloop}). The system is supported by carefully designed reward structures and randomization techniques (\cref{sec:reward}) and trained on a newly curated dataset, \dataset, that ensures robust generalization (\cref{sec:data}). Additional details of the implementation are provided in \cref{sec:supp:impl}.

\subsection{Policy Learning}\label{sec:training}

We employ a teacher-student training strategy for the teleoperation policy, following the overall framework of OmniH2O~\cite{he2024omnih2o} (see \cref{sec:supp:formulation} for the problem formulation). This approach first trains a teacher policy with privileged information, then distills this knowledge into a student policy that operates using only real-world observations.

\paragraph{Teacher Policy Training}

The teacher policy $\pi_{\text{tea}}$ is implemented as an \ac{mlp} that leverages comprehensive state information unavailable on real robots. At each timestep $t$, it processes observations $\mathbf{o}_{t}^{\text{tea}} = [\mathbf{s}^{\text{tea}}_{t}, \mathbf{o}^{\text{task}}_{t}, \mathbf{a}_{t}, \mathbf{o}^{\text{env}}_{t}]$ and outputs target joint positions $\mathbf{a}_{t+1} \in \mathbb{R}^{29}$ for PD control. The privileged states $\mathbf{s}^{\text{tea}}_{t} = [\mathbf{p}_{t}, \mathbf{\theta}_{t}, \mathbf{v}_{t}, \mathbf{\omega}_{t}]$ include joint angular positions $\mathbf{p}_{t}$ and the 6D poses, linear velocities, and angular velocities $\mathbf{\theta}_{t}, \mathbf{v}_{t}, \mathbf{\omega}_{t}$ of all robot links. Task observations $\mathbf{o}^{\text{task}}_{t} = [\hat{\mathbf{p}}_{t+1} - \mathbf{p}_{t}, \hat{\mathbf{\theta}}_{t+1} - \mathbf{\theta}_{t}, \hat{\mathbf{v}}_{t+1} - \mathbf{v}_{t}, \hat{\mathbf{\omega}}_{t+1} - \mathbf{\omega}_{t}, \hat{\mathbf{p}}_{t+1}, \hat{\mathbf{\theta}}_{t+1}]$ capture both reference motion (denoted by $\hat{\cdot}$) and tracking errors between reference and current states. Environmental observations $\mathbf{o}^{\text{env}}_t$ provide context including ground friction coefficient and robot mass distribution.

\paragraph{Student Policy Distillation}

The student policy must operate without privileged information, following $a_{t+1} = \pi_{\text{stu}}(s^\text{stu}_{t-25:t}, a_{t-25:t}, o^\text{task}_t)$. The robot state sequence $s^{stu}_{t-25:t}$ contains joint positions $q$, joint velocities $\dot{q}$, root angular velocity $\omega^{root}$, and root gravity vector $g$ obtained from on-device IMU over the past 25 frames. Task observations $o^{task}_t$ consist of $\hat{p}_{t+1} - p_t$, $\hat{p}_{t+1}$, $\hat{\dot{p}}_{t+1}$, $h_{t}$, and $\hat{h}_{t+1}$, where $p_t$ represents the 3D positions of head and two wrists obtained from LiDAR odometry and forward kinematics, $\hat{p}_{t+1}$ and $\hat{\dot{p}}_{t+1}$ are target positions and velocities from reference motion, and $h_t$, $\hat{h}_{t+1}$ represent current and target wrist orientations.

The key challenge lies in handling diverse motion patterns within a single policy. Walking requires different control strategies than crouching or reaching, yet traditional monolithic architectures struggle with these conflicting objectives. We address this through an \ac{moe} architecture, as shown in \cref{fig:model}, which allows specialized processing for different motion types.

The \ac{moe} design consists of $L$ layers, each comprising $N$ experts that function as independent feed-forward sub-layers with distinct parameters. At each layer, a router dynamically selects which experts are activated based on the input, generating weight distributions over all experts. The layer output combines the top-$k$ experts with highest routing weights: $f = \sum_{i}^{k}{w_i\cdot E_i(\cdot)}$, where $w_i$ is the routing weight for the $i$-th selected expert and $E_i(\cdot)$ is the output of the $i$-th expert. This design enables different experts to focus on distinct motion patterns. To prevent model collapse to only a few experts, we introduce a balancing loss that encourages uniform expert selection:
\begin{small}
\begin{equation}
    \mathcal{L}_{balance}=\sum_{l=1}^L\sum_{e=1}^N[\max (p_e - \frac{1+\epsilon}{N}, 0)+\min (\frac{1-\epsilon}{N}-p_e, 0)],
\end{equation}
\end{small}
where $p_e = \mathbb{E}[w_e]$ represents the expected activation probability of expert $e$, and $\epsilon$ is a slack constant that allows slight deviations from perfect uniformity.

\subsection{Closed-Loop Error Correction}\label{sec:closedloop}

Traditional humanoid teleoperation systems operate in an open-loop configuration, where small errors in position tracking accumulate over time, leading to significant drift during extended operations. This fundamental limitation becomes particularly problematic during long-horizon tasks that require sustained positional accuracy. To address this challenge, we implement a closed-loop error correction mechanism that continuously monitors and compensates for positional discrepancies between the teleoperator and the humanoid.

Our approach utilizes LiDAR odometry to maintain accurate global position estimates for both the humanoid and the teleoperator. We employ FAST-LIO2~\cite{xu2022fast}, an algorithm that tightly couples IMU and LiDAR data through an iterated Kalman filter to provide robust real-time state estimation even during dynamic movements (see more details in \cref{sec:supp:odometry}). This choice ensures reliable tracking performance across diverse motion patterns, from walking to complex manipulation tasks.

The system tracks global positions for both agents: the humanoid's position $p \in \mathbb{R}^3$ is computed from onboard sensors, while the teleoperator's position $\hat{p} \in \mathbb{R}^3$ is similarly tracked through a \ac{mr} hardware equipped with a comparable odometry pipeline. The student teleoperation policy directly consumes the difference between $p$ and $\hat{p}$, enabling it to generate actions that systematically reduce positional drift and maintain accurate correspondence between the operator and the humanoid.

\subsection{Reward Design and Domain Randomization}\label{sec:reward}

We build upon the reward terms and domain randomizations from OmniH2O~\cite{he2024omnih2o} as the foundation of our approach, with specific enhancements to address the challenges of real-world teleoperation. Detailed reward functions and domain randomization settings are provided in \cref{sec:supp:reward}.

To enhance robustness against LiDAR odometry errors, we introduce a velocity-dependent \ac{sde} noise model during training that reflects real-world error characteristics. For the head position $\Vec{p}_{\mathrm{head}}$, we define the randomized position $\Vec{P}_{\mathrm{head}}$ as:
\begin{small}
\begin{equation}
    \mathrm{d}\Vec{P}_\mathrm{head} = \Dot{\Vec{p}}_\mathrm{head}\mathrm{d}t + (\frac{\parallel\Dot{\Vec{p}}_\mathrm{head}\parallel}{c_\mathrm{vel}}+c_\mathrm{min})\mathrm{d}\Vec{W},
\end{equation}
\end{small}
where $\vec{W}$ is a standard Wiener process, and $c_{\mathrm{vel}}$ and $c_{\mathrm{min}}$ are constants that scale the noise proportionally to movement speed and establish a minimum randomization level. This formulation mirrors real-world dynamics, where faster movements tend to produce greater odometry errors. We use forward kinematics to compute other body positions based on the randomized head position, while periodically resetting and constraining the maximum deviation to avoid unrealistic drift.

Since \model provides only upper-body references (head and wrists), we must generate appropriate lower-body behaviors without explicit guidance. To tackle this challenge, we employ an \ac{amp} reward~\cite{peng2021amp} to regularize lower-body movements and encourage natural, stable behavior. Through this combination of specialized domain randomization and reward design, \model learns to generate robust lower-body behaviors while maintaining precise upper-body control aligned with operator commands.

\subsection{The \dataset Dataset}\label{sec:data}

The training dataset \dataset comprises three complementary components to support robust whole-body teleoperation. These include: (i) an augmented AMASS~\cite{mahmood2019amass} subset of 149 curated sequences featuring diverse pairings of upper- and lower-body movements, enhanced via motion editing to increase compositional diversity and policy generalization(see \cref{sec:supp:data}); (ii) 14 custom sequences captured with an IMU-based Xsens \ac{mocap} system to fill coverage gaps, emphasizing continuous transitions and diverse upper-body poses critical for manipulation; and (iii) systematic hand orientation augmentation through procedurally generated 6D wrist targets, smoothed via \ac{slerp} to ensure coherent and natural hand motions for teleoperation.

\section{Real-World Experiments}

\begin{wrapfigure}{r}{0.4\linewidth}
    \centering
    \small
    \vspace{-39pt}
    \includegraphics[width=\linewidth]{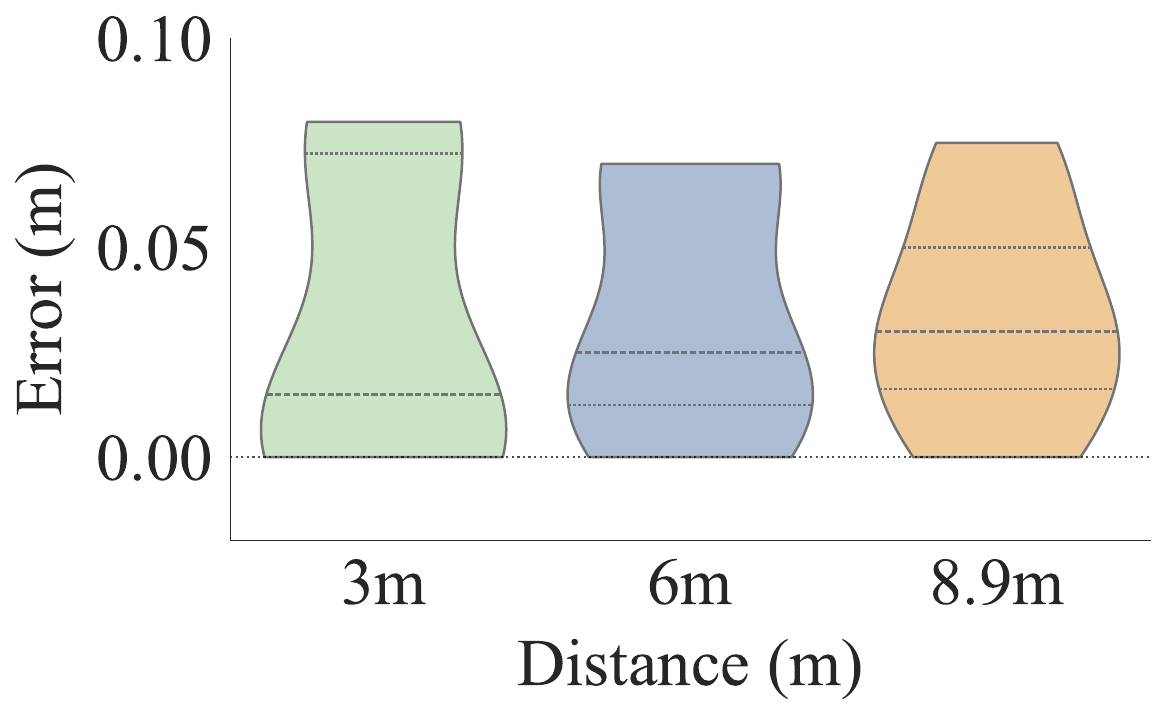}
    \caption{\textbf{Global position tracking accuracy in real-world experiments.} \model achieves mean tracking errors of 5.1cm across distances up to 8.9m, demonstrating effective closed-loop error correction in extended teleoperation.}
    \label{fig:real}
    \vspace{-12pt}
\end{wrapfigure}

We evaluated \model on a physical Unitree G1 humanoid through comprehensive experiments demonstrating exceptional whole-body motion fidelity and precise position tracking. Our experiments focus on two key capabilities: (i) global position tracking accuracy during extended teleoperation to validate our closed-loop error correction mechanism, and (ii) whole-body motion tracking fidelity across diverse skills to demonstrate coordination capabilities. These experiments collectively validate both the technical performance and practical applicability of our approach to real-world humanoid teleoperation.

\paragraph{Global Position Tracking}
To evaluate global positioning accuracy over extended distances, we designed a controlled path-following experiment.
(i) \textbf{Straight-path tracking.} We fixed the initial positions of both the operator and the robot. The operator then walked along straight paths to target positions at $3$m, $6$m, and $8.9$m while teleoperating the robot. We measured the discrepancy between the robot’s final and expected positions as the tracking error and repeated each condition ten times.
(ii) \textbf{Curved-path tracking.} We teleoperated the humanoid along a $10$m trajectory with two $90^{\circ}$ turns, representative of typical household paths, and repeated the trial six times to measure translational and rotational tracking errors.

\model achieved a mean tracking error of $5.1$ cm in straight-path tracking, with a maximum deviation of $12.0$ cm at $8.9$m (see \cref{fig:real}). These results show that \model's closed-loop error correction effectively mitigates accumulated errors during extended teleoperation.

Statistical analysis confirms consistent tracking performance across all tested distances. Independent samples t-tests comparing distance groups yielded: $3$m versus $6$m ($t=0.165, p=0.871$) and $6$m versus $8.9$m ($t=0.048, p=0.963$). With all p-values $> 0.05$, \model shows no significant performance degradation across the tested range, demonstrating that our approach maintains high accuracy over extended distances without drift accumulation.

In curved-path tracking, the mean error is $20$ cm (maximum $27$ cm), and the mean rotational drift was $2^{\circ}$ between the operator’s and humanoid’s orientations. These results highlight \model’s ability to sustain high precision even over long and complex trajectories.




\paragraph{Whole-Body Motion Tracking}

\begin{figure*}[t!]
    \centering
    \begin{subfigure}[b]{0.42\linewidth}
        \includegraphics[width=\linewidth]{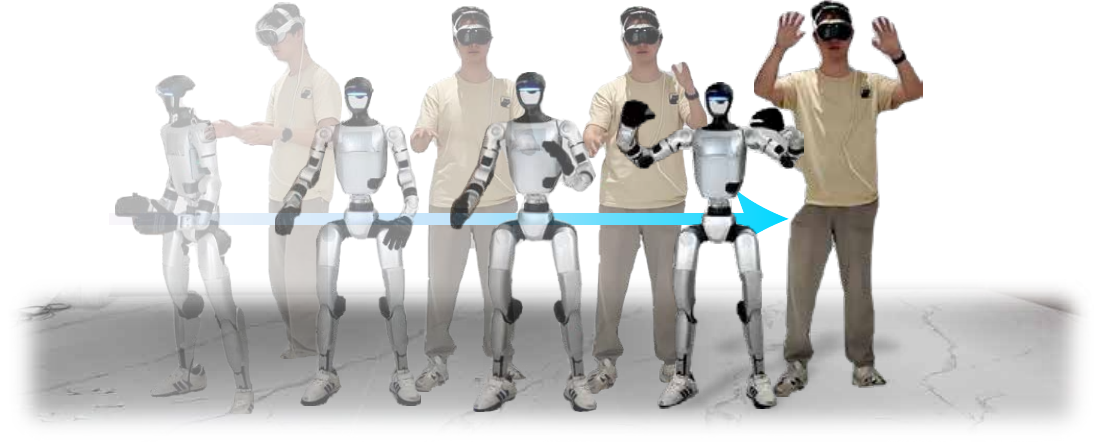}
        \caption{Waving}
    \end{subfigure}%
    \begin{subfigure}[b]{0.58\linewidth}
        \includegraphics[width=\linewidth]{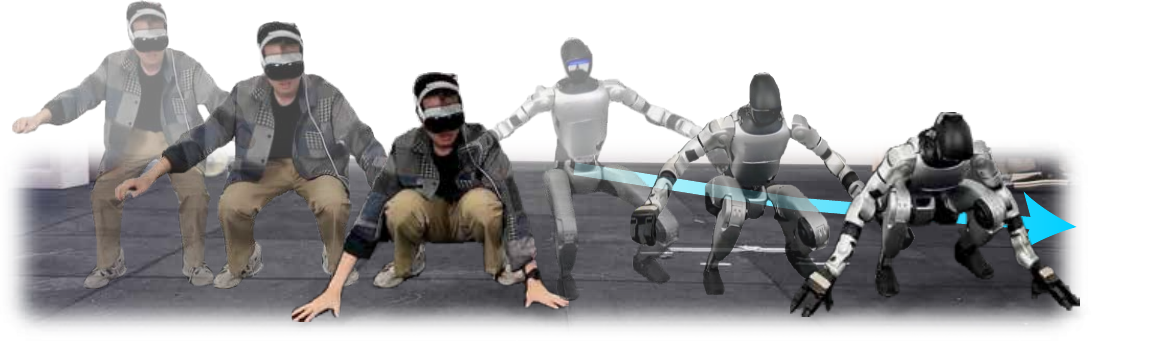}
        \caption{Squatting}
    \end{subfigure}%
    \\%
    \begin{subfigure}[b]{0.515\linewidth}
        \includegraphics[width=\linewidth]{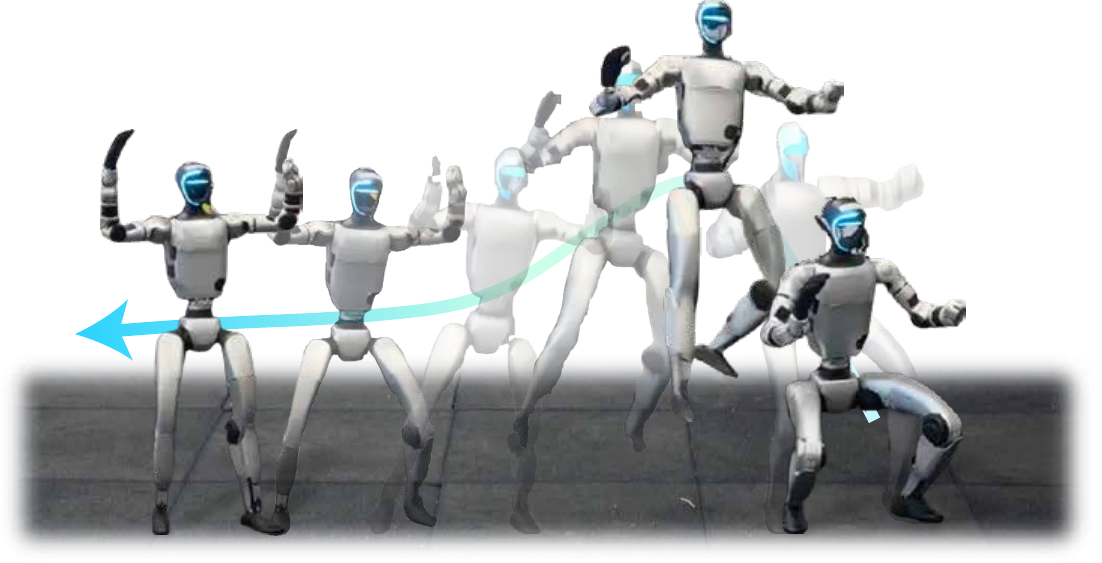}
        \caption{Jumping}
    \end{subfigure}%
    \begin{subfigure}[b]{0.485\linewidth}
        \includegraphics[width=\linewidth]{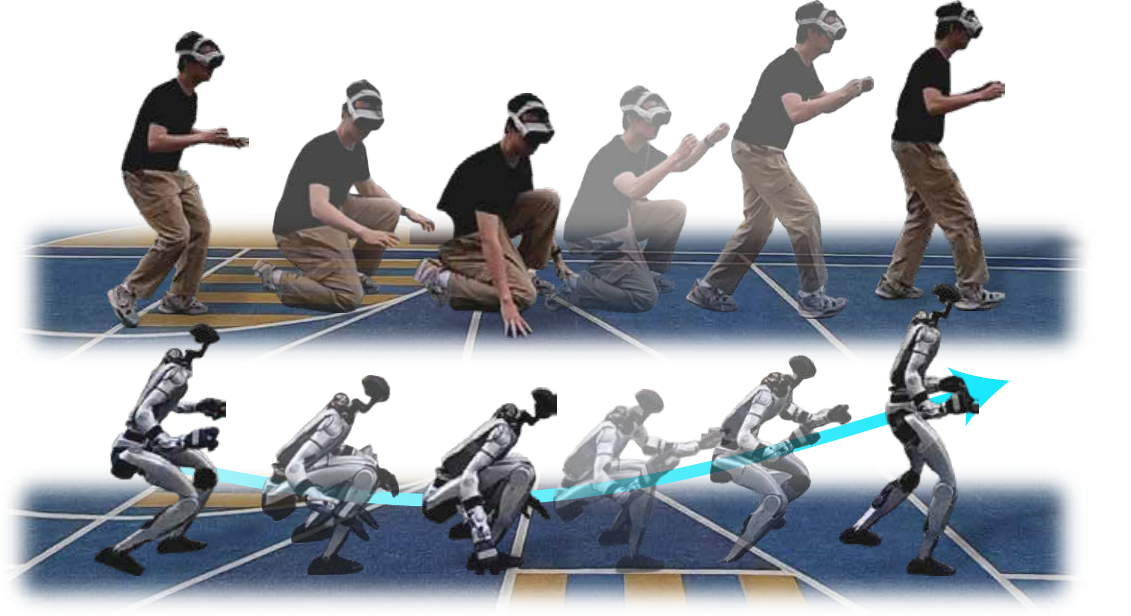}
        \caption{Squatting}
    \end{subfigure}%
    \caption{\textbf{Whole-body motion tracking on Unitree G1.} \model successfully tracks diverse skills including (a) waving, (b)(d) squatting, and (c)jumping, showcasing comprehensive whole-body coordination capabilities.}
    \label{fig:tracking}
\end{figure*}

As shown in \cref{fig:tracking}, \model successfully enables real-time teleoperation across a diverse range of whole-body skills. The robot accurately tracks complex motions including waving, squatting, standing up from squatted positions, and jumping. These results demonstrate high whole-body motion fidelity for real-time humanoid teleoperation, particularly for dynamic skills like jumping requiring precise coordination of balance control and force application.

\begin{figure*}[b!]
    \centering
    \begin{subfigure}[b]{\linewidth}
        \includegraphics[width=.25\linewidth]{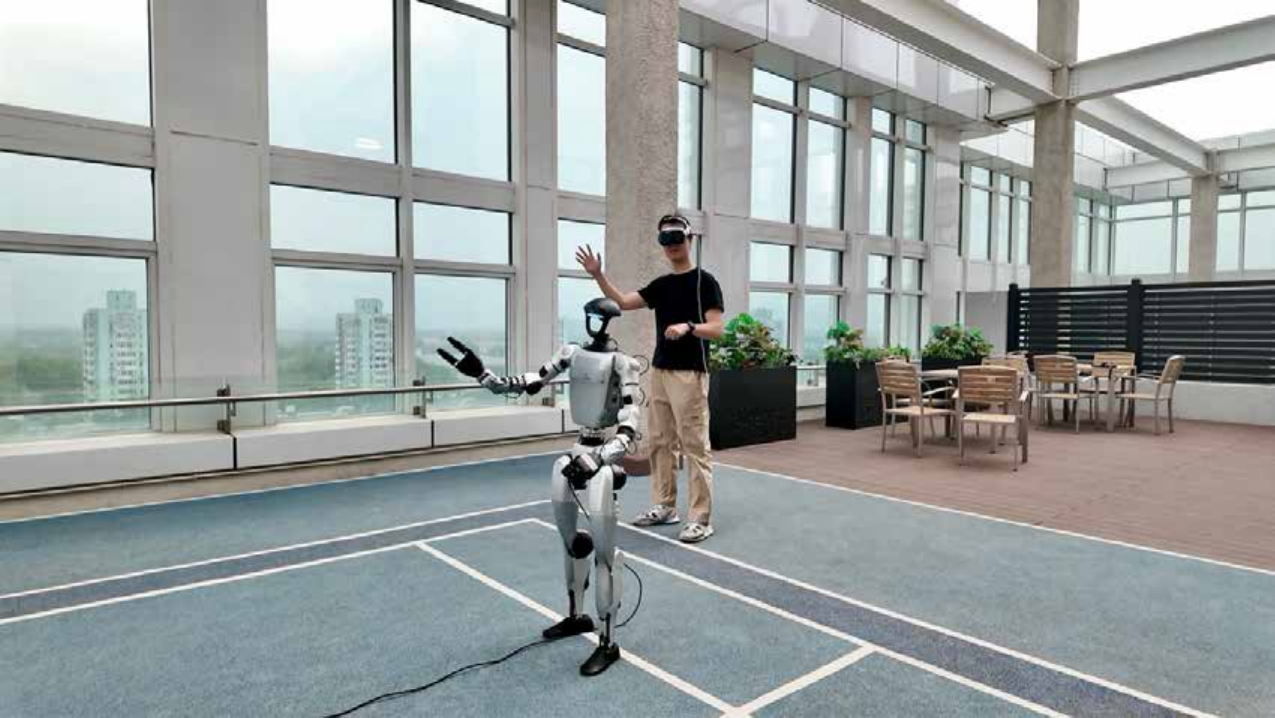}%
        \includegraphics[width=.25\linewidth]{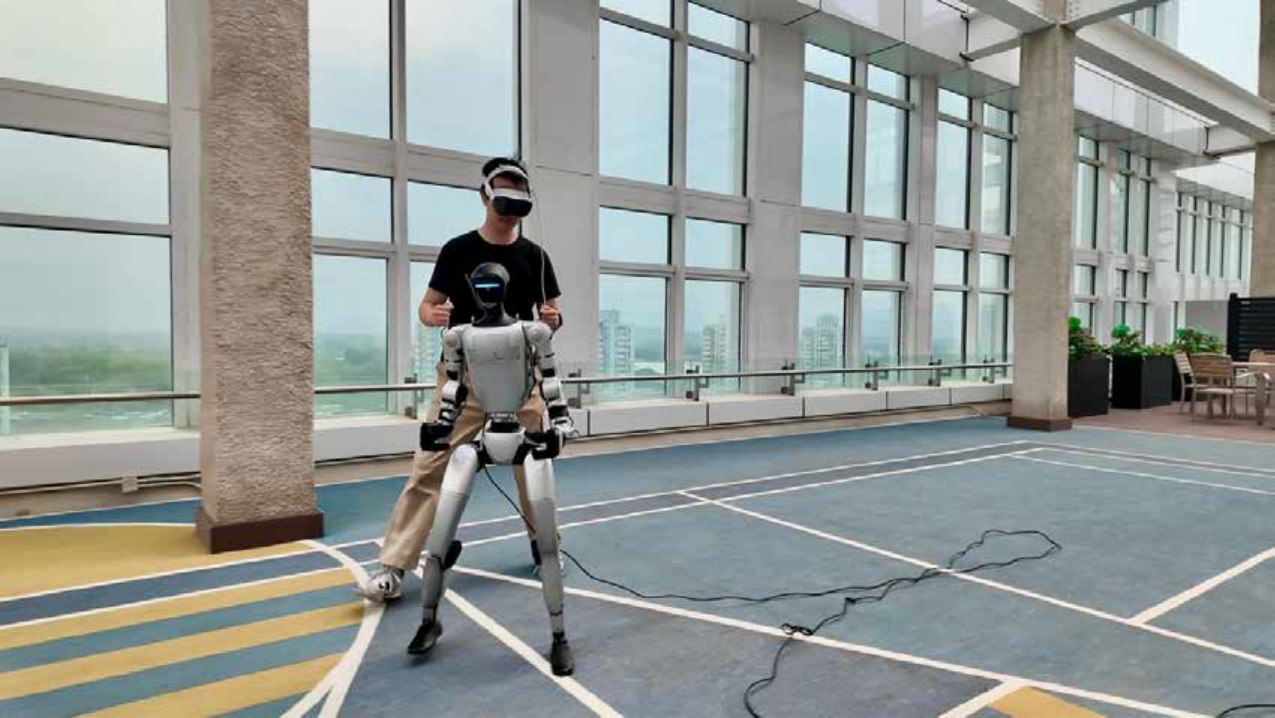}%
        \includegraphics[width=.25\linewidth]{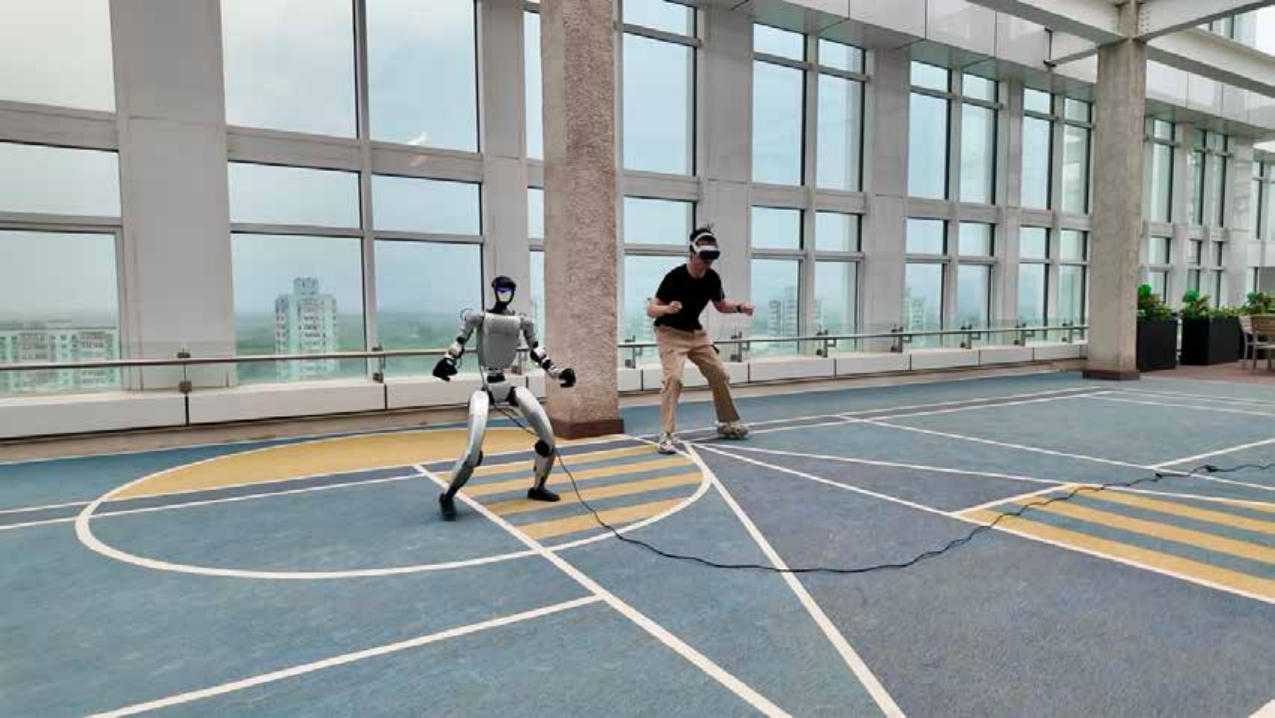}%
        \includegraphics[width=.25\linewidth]{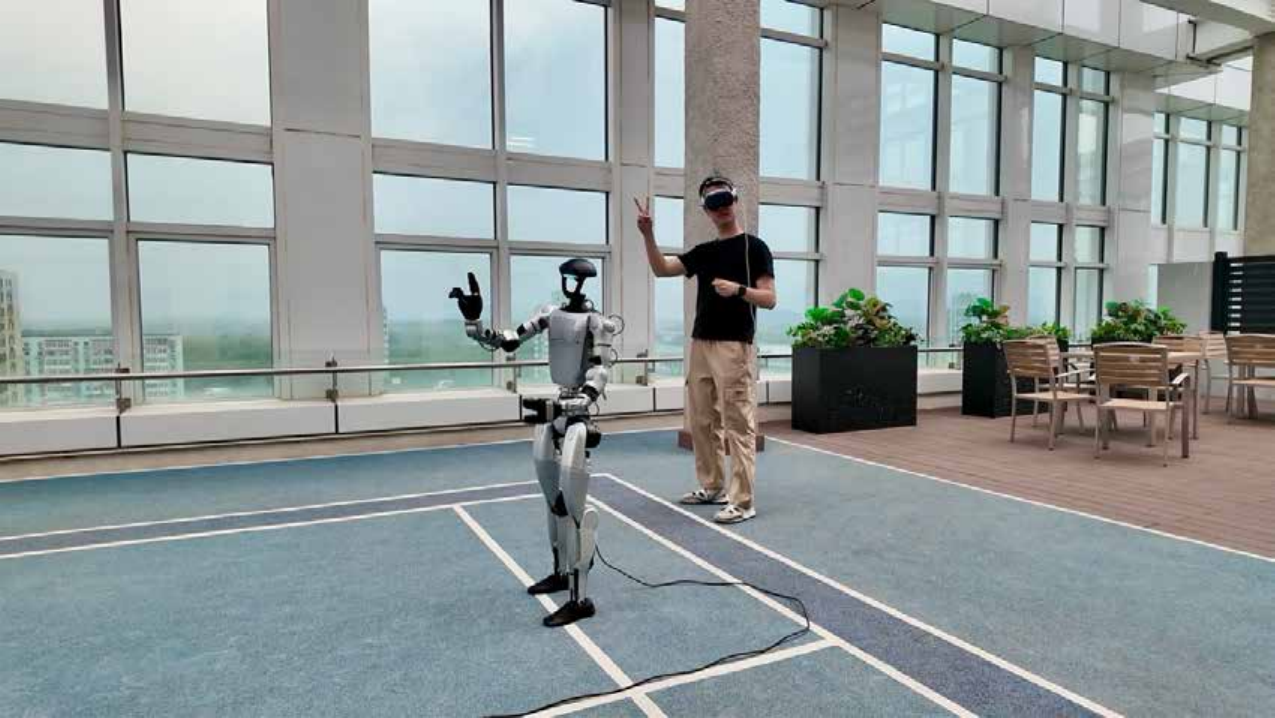}%
        \caption{Sequence 1}
    \end{subfigure}%
    \\%
    \begin{subfigure}[b]{\linewidth}
        \includegraphics[width=.25\linewidth]{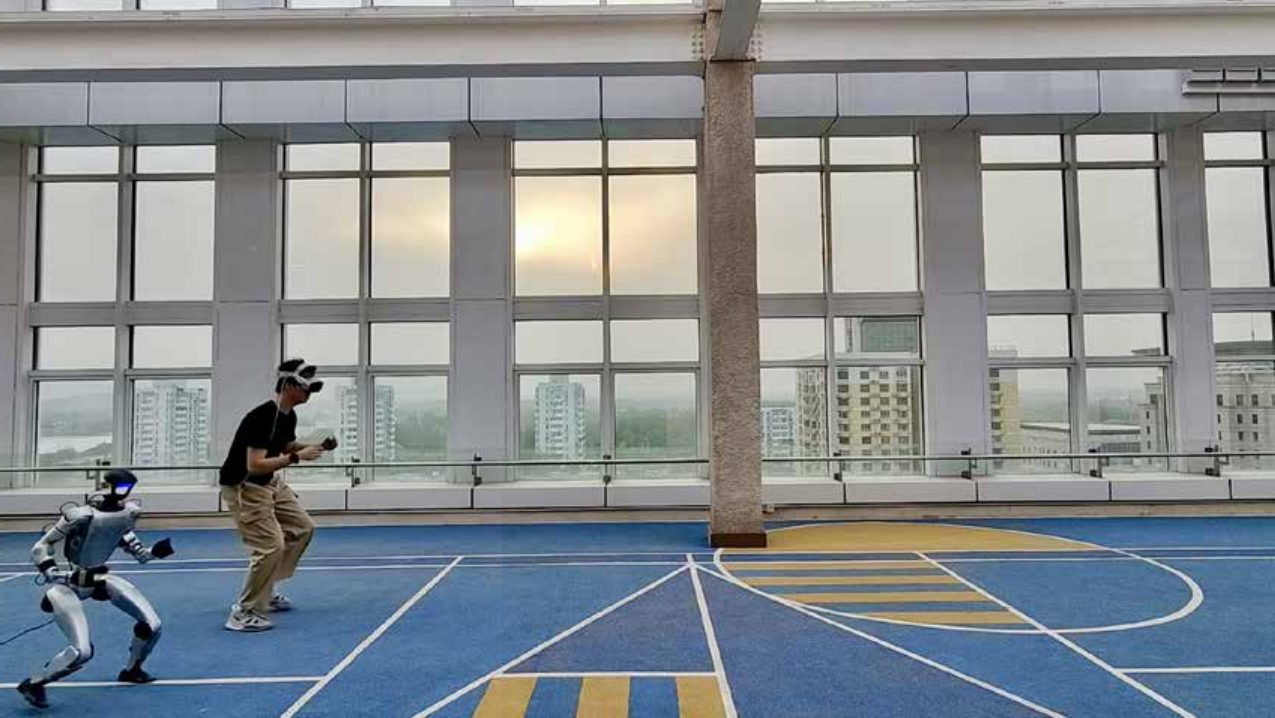}%
        \includegraphics[width=.25\linewidth]{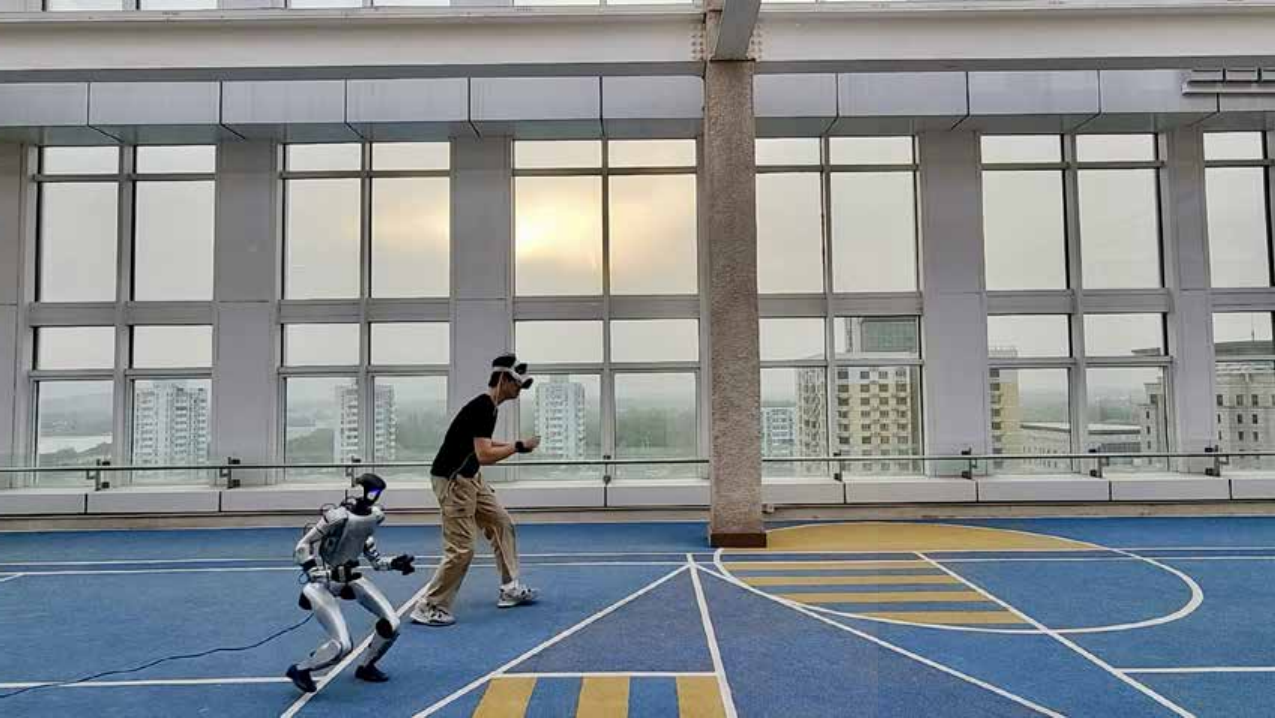}%
        \includegraphics[width=.25\linewidth]{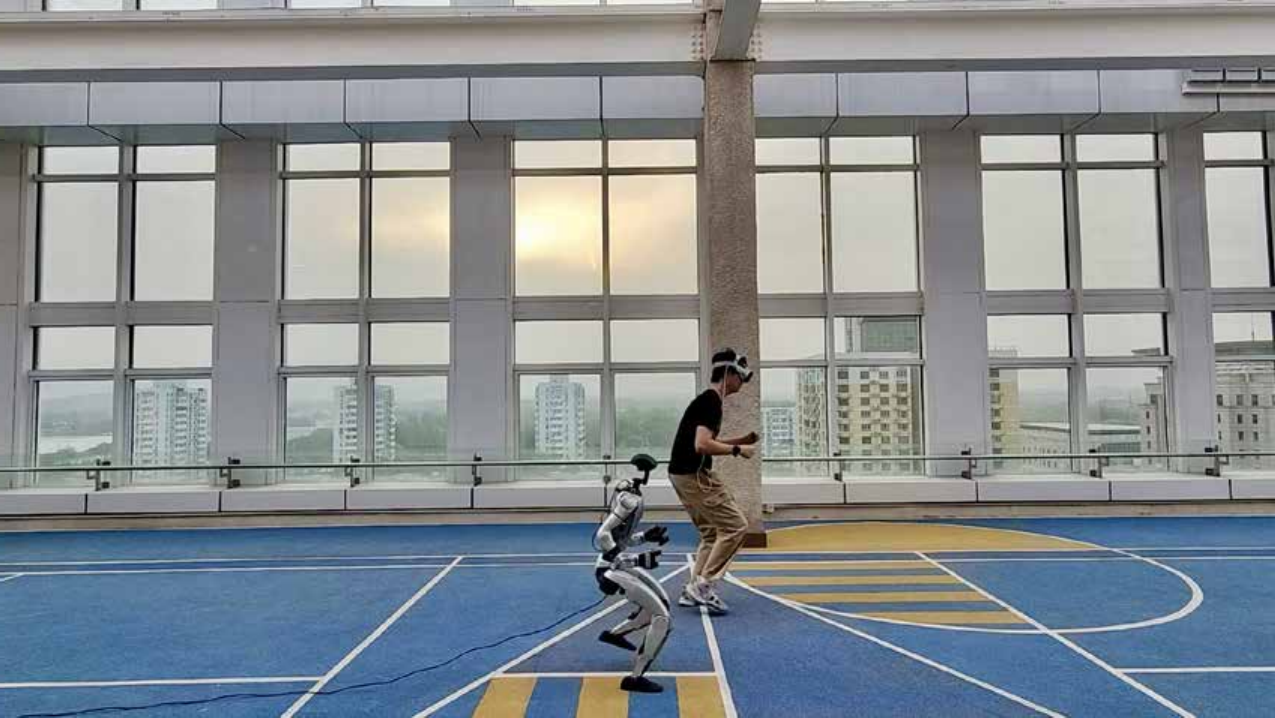}%
        \includegraphics[width=.25\linewidth]{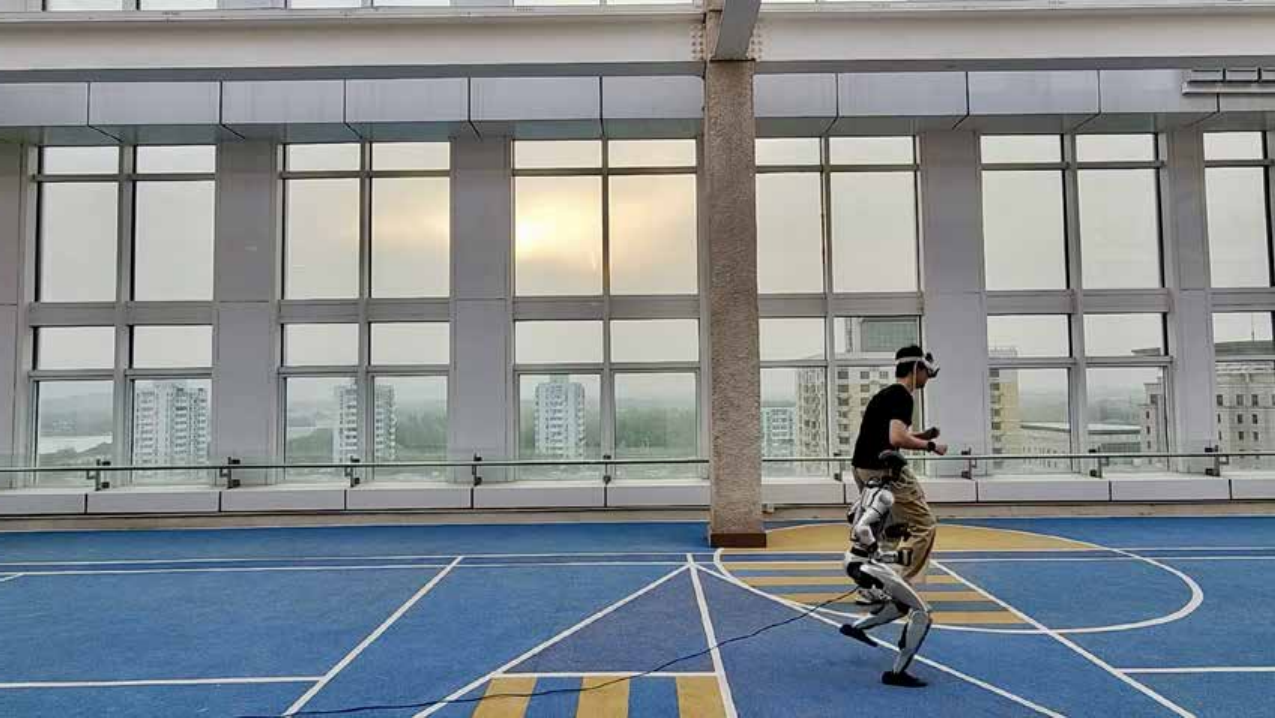}%
        \caption{Sequence 2}
    \end{subfigure}%
    \caption{\textbf{Long-horizon teleoperation.} The humanoid accurately tracks both the operator's local pose and global translation throughout a complex navigation sequence, demonstrating robust performance.}
    \label{fig:long_tele}
\end{figure*}

\paragraph{Long-Horizon Mixed Navigation}

To validate system performance in complex scenarios, we conducted extended teleoperation sessions incorporating multiple movement types. As visualized in \cref{fig:long_tele}, we recorded a continuous teleoperation sequence where the operator traversed a complex path spanning over $15$m, incorporating diverse locomotion patterns including forward walking, turning, side-stepping, and returning to the original position.

The robot consistently tracked the operator's movements throughout this extended sequence and returned to its starting position with minimal drift. This result validates \model's robustness for extended teleoperation sessions that combine locomotion and whole-body motion control---a critical capability for practical humanoid applications requiring sustained coordination over long horizons.

\section{Simulations}

We present comprehensive evaluations of \model in simulation across four key settings: reference motion tracking, diverse stance tracking, ablation studies (\cref{sec:supp:ablation}), and expert activation analysis (\cref{sec:supp:moe}). These experiments are designed to: (i) quantify motion tracking accuracy in controlled simulation environments, and (ii) assess robustness across diverse stance configurations. The evaluation metrics are detailed in \cref{sec:supp:metric}.

\begin{wraptable}{r}{0.5\linewidth}
    \scriptsize
    \centering
    \setlength{\tabcolsep}{3pt}
    \vspace{-10pt}
    \caption{\textbf{Motion tracking evaluation on \dataset dataset.} Comparison of \model against ablations: \model\textsuperscript{\(\dagger\)} uses an \acs{mlp} instead of \acs{moe} architecture, while \model\textsuperscript{\(*\)} trains on OmniH2O data instead of \dataset.}
    \label{tab:baseline}
    \begin{tabular}{@{}l *{5}{c}@{}}
        \toprule
        \textbf{Method} & 
        \textbf{$\mathbf{SR}$} $\uparrow$ & 
        \textbf{$E_{\text{mpkpe}}$} $\downarrow$ & 
        \textbf{$E_{\text{r-mpkpe}}$} $\downarrow$ & 
        \textbf{$E_{\text{vel}}$} $\downarrow$ & 
        \textbf{$E_{\text{hand-rot}}$} $\downarrow$ \\
        \midrule
        \model\textsuperscript{\(\dagger\)} & $100\%$ & $113.97$ & $35.55$ & $245.11$ & $4.73$\\
        \model\textsuperscript{*} & $100\%$ & $102.20$ & $41.07$ & $309.65$ & $4.61$\\
        \model & $100\%$ & $\textbf{87.84}$ & $\textbf{33.30}$ & $\textbf{227.17}$ & $\textbf{3.61}$\\
        \bottomrule 
    \end{tabular}
\end{wraptable}

\paragraph{Motion Tracking}

We compared \model against two ablated baselines: \model\textsuperscript{\(\dagger\)} and \model\textsuperscript{\(*\)}. \model\textsuperscript{\(\dagger\)} employs an \ac{mlp} as the student policy, resembling the OmniH2O baseline trained on our data and task. \model\textsuperscript{\(*\)} represents our \model trained on OmniH2O data instead of \dataset. Quantitative results in \cref{tab:baseline} reveal that both the \ac{moe} architecture and \dataset contribute significantly to accurate reference motion tracking. Qualitative comparisons are provided in \cref{sec:supp:compare}.

\paragraph{Tracking Diverse Stances}

\begin{wrapfigure}{r}{0.5\linewidth}
    \centering
    \small
    \vspace{-18pt}
    \includegraphics[width=.5\linewidth]{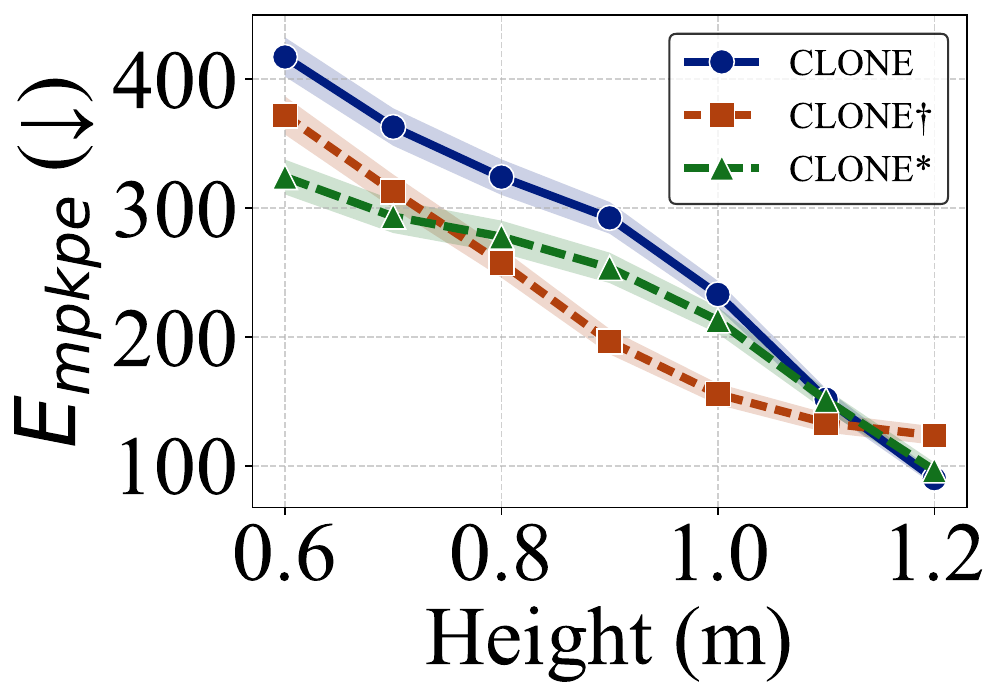}%
    \includegraphics[width=.5\linewidth]{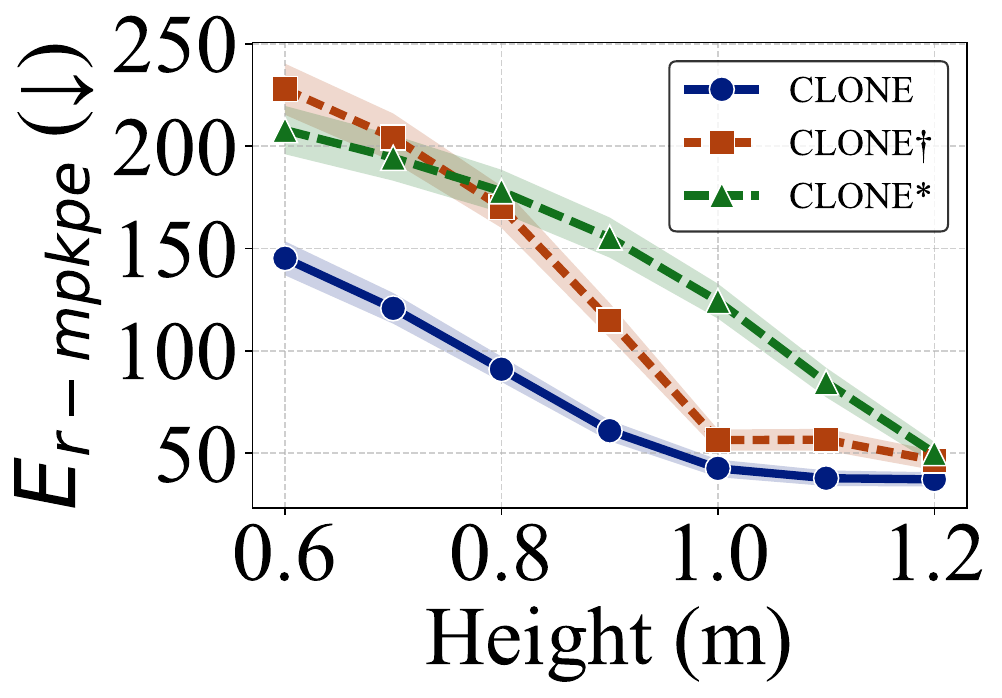}%
    \\%
    \includegraphics[width=.5\linewidth]{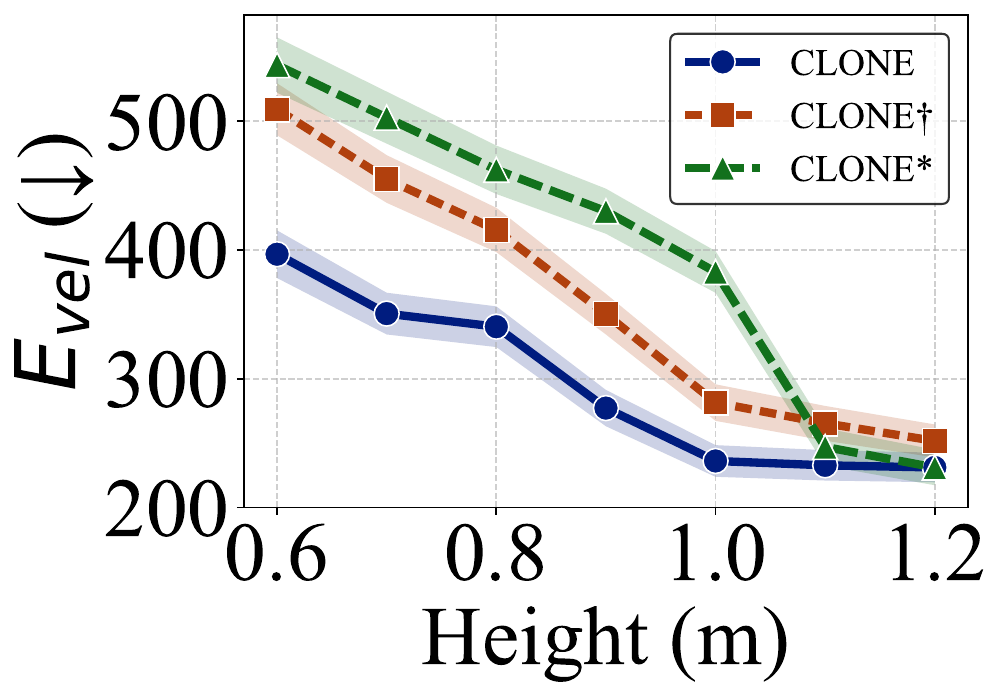}%
    \includegraphics[width=.5\linewidth]{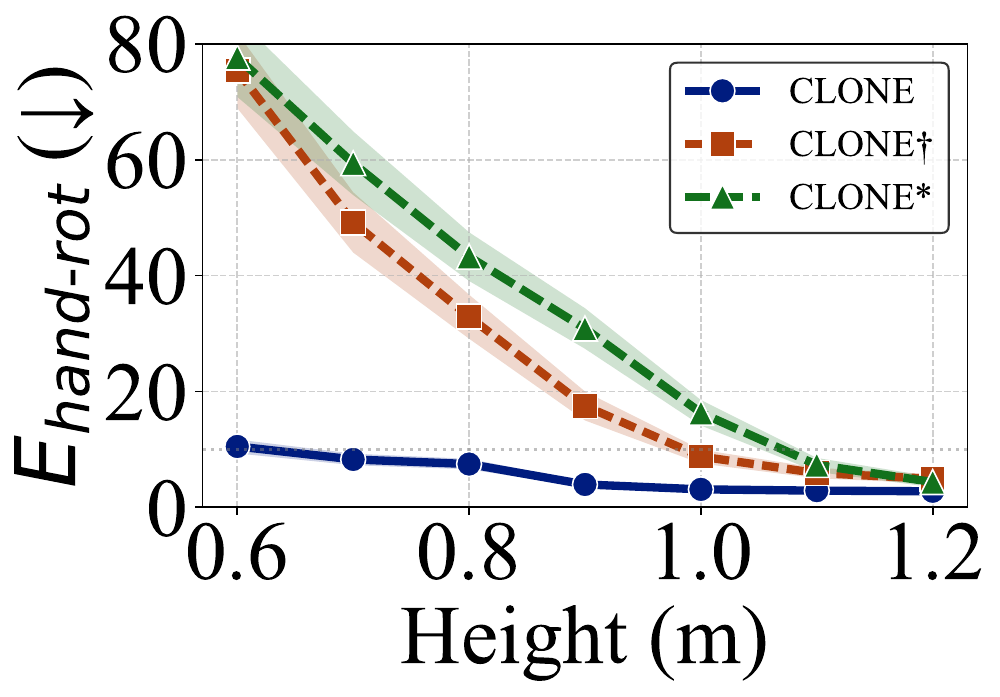}%
    \caption{\textbf{Motion tracking performance across stance heights.} Comparison between \model (blue solid), \model\textsuperscript{$*$} (green dashed), and \model\textsuperscript{$\dagger$} (red dashed) across different postures from standing to deep squatting. Lower values indicate better performance.}
    \label{fig:heights}
    \vspace{-12pt}
\end{wrapfigure}

To assess \model's robustness across varying postures, we evaluated performance in tracking motions with head heights from $1.2$m (standing) to $0.6$m (deep squatting) in $0.1$m decrements. We generated these challenging motions by systematically editing sequences from the \dataset dataset, creating unseen poses that test teleoperation system limits.

\cref{fig:heights} reveals an interesting performance trade-off: while \model underperforms baselines in absolute position accuracy (MPKPE), it consistently outperforms them in local motion metrics (R-MPKPE, velocity error, and hand orientation). This pattern indicates that \model prioritizes faithful reproduction of reference motions---particularly for challenging postures---sometimes at the expense of global positioning accuracy. All methods exhibit increased tracking errors at lower heights, confirming the inherent difficulty of teleoperating robots in squatting postures.

\section{Conclusion}
We present \model, a closed-loop \acs{moe}-based teleoperation system that enables comprehensive humanoid control while addressing accumulated tracking errors in long-horizon tasks. Our approach integrates three key innovations: (i) an \acs{moe} architecture that coordinates diverse motion skills while maintaining natural upper- and lower-body coordination, (ii) LiDAR-based closed-loop error correction that prevents positional drift accumulation through real-time feedback, (iii) and the \dataset dataset that augments AMASS with hand orientations and additional motion-captured sequences for robust whole-body coordination training.
Experimental validation demonstrates exceptional performance: $5.1$ cm mean global position tracking error over $8.9$ m trajectories, accurate whole-body coordination across diverse skills including complex coordinated movements like object retrieval from ground level, and robust long-horizon teleoperation requiring only head and hand tracking from a single commercial MR headset.

\section{Limitations}
While \model demonstrates significant capabilities, it still has important limitations that warrant further investigation. The minimal input configuration, though enabling intuitive control with head and hand tracking, inherently limits fine-grained stability control in certain scenarios. Additionally, the system exhibits reduced performance during highly dynamic movements like jumping, stemming from training data constraints and balance control challenges.

Future work should explore additional sensing modalities to enhance stability while preserving interface simplicity, and expand motion datasets with specialized reward functions for dynamic behaviors. \model establishes new benchmarks for practical humanoid teleoperation, achieving unprecedented fidelity in complex tasks while maintaining essential simplicity.







\section*{Acknowledgments}

The authors gratefully acknowledge Unitree Robotics for their support with hardware. This work is supported in part by the National Science and Technology Major Project (2022ZD0114900), the National Natural Science Foundation of China (62376031), the Beijing Nova Program, the State Key Lab of General AI at Peking University, the PKU-BingJi Joint Laboratory for Artificial Intelligence, and the National Comprehensive Experimental Base for Governance of Intelligent Society, Wuhan East Lake High-Tech Development Zone.

\bibliography{reference_header,reference}

\clearpage
\appendix
\renewcommand\thefigure{A\arabic{figure}}
\setcounter{figure}{0}
\renewcommand\thetable{A\arabic{table}}
\setcounter{table}{0}
\renewcommand\theequation{A\arabic{equation}}
\setcounter{equation}{0}
\pagenumbering{arabic}
\renewcommand*{\thepage}{A\arabic{page}}
\setcounter{footnote}{0}

\section{Preliminaries}

\subsection{Formulation}\label{sec:supp:formulation}

We formulate the humanoid teloperation as a \ac{mdp} $\mathcal{M} =\{\mathcal{S}, \mathcal{A}, \mathcal{T}, \mathcal{R}\}$. $\mathcal{S}$ includes proprioceptive states~$s$ and task-oriented observations~$o^{task}$. The action space $\mathcal{A} \in \mathcal{R}^{29}$ represents the humanoid's joint angles in our method. $\mathcal{T}$ is the transition function conditioned on $\mathcal{S}$ and $\mathcal{A}$. The reward functions $\mathcal{R}$ is defined based on $\mathcal{S}, \mathcal{A}$. A policy $\pi$ is proposed to maximize the overall reward $\mathcal{R}$ using the \ac{ppo} algorithm. 

\subsection{LiDAR Odometry and Closed-Loop Error Correction}\label{sec:supp:odometry}

LiDAR odometry is designed to accurately determine the robot's state, including its orientation and position. In this paper, we adopt FAST-LIO2 \cite{xu2022fast}, which utilizes onboard LiDAR and IMU to estimate the humanoid's global position. FAST-LIO2 \cite{xu2022fast} leverages IMU data and LiDAR point clouds to construct and update a 3D map in real time. It then registers the current LiDAR point clouds with the map to estimate the robot's current position.

Previous teleoperation systems \cite{he2024omnih2o,cheng2024expressive,ji2024exbody2} often operate in an open-loop manner, primarily due to the absence of the humanoid's global position. Consequently, stepwise tracking errors accumulate over time, leading to significant drift during long-horizon tasks. In this work, we leverage LiDAR odometry to determine the robot's global position. Similarly, we obtain the operator's global position with the odometry from Apple Vision Pro. We integrate the difference between the two positions into the task observation $o^{task}$, and design a reward function for our teleoperation policy that minimizes this difference. Notably, the LiDAR operates at 10 Hz, and our policy runs at 50 Hz. Our policy uses the latest available odometry position at each timestep.

\section{Reward Functions and Domain Randomization}\label{sec:supp:reward}

\cref{tab:supp:reward} provides a detailed overview of the reward structure utilized in this study, while \cref{tab:dr} outlines the domain randomization scheme employed.

\begin{table}[ht!]
    \centering
    \small
    \caption{\textbf{Reward functions.} The details of the primary reward function used in our training process.}
    \label{tab:supp:reward}
    \setlength{\tabcolsep}{2pt}
    \begin{tabular}{ccc}
        \toprule
        Term  & Expression & Weight \\
        \midrule
        Torque & $\lVert \boldsymbol{\tau} \rVert_2^2$ & $-0.0001$ \\ 
        Torque limits & $ [{\tau \notin [{\tau}_{\text{min}}, {\tau}_{\text{max}} ]}]_1$ & -2 \\
        DoF position limits & $ [{\mathbf{p}_t \notin [\mathbf{p}_{\text{min}}, \mathbf{p}_{\text{max}} ]}]_1 $ & -625\\
        DoF velocity limits & $ [{\mathbf{\dot{p}}_t \notin [\mathbf{\dot{p}}_{\text{min}}, \mathbf{\dot{p}}_{\text{max}} ]}]_1 $ & -50\\
        Termination & $\text{termination}_1$ & $-e^4$ \\ 
        DoF acceleration & $\lVert \mathbf{\ddot{q}_t} \rVert_2^2$ & $-1.1e^{-5}$ \\  
        DoF velocity & $\lVert \mathbf{\dot{q}_t} \rVert_2^2$ & $-0.004$ \\  
        Lower-body action rate & $\lVert \mathbf{a}_t^{\text{lower}} - \mathbf{a}_{t-1}^{\text{lower}} \rVert_2^2$ & $-1.0$ \\  
        Upper-body action rate & $\lVert \mathbf{a}_t^{\text{upper}} - \mathbf{a}_{t-1}^{\text{upper}} \rVert_2^2$ & $-0.3$ \\  
        Feet air time & $T_{\text{air}} - 0.3$ & $2500$ \\
        Stumble & $[(\mathbf{F}_{\text{feet}}^{xy} > 5\times\mathbf{F}_{\text{feet}}^z)]_1$ & $-1.25e^{4}$ \\  
        Slippage & $\lVert v^\text{feet}_t \rVert_2^2 \cdot [(\mathbf{F}_{\text{feet}} \geq 1)]_1$ & $-80$ \\  
        Feet orientation & $\lVert \mathbf{R}_z^{\text{feet}} \rVert$ & $-62.5$ \\  
        In the air & $[(\mathbf{F}_{\text{feet}}^{\text{left}}, \mathbf{F}_{\text{feet}}^{\text{right}} < 1)]_1$ & $-750$ \\  
        Orientation & $\lVert \mathbf{R}_z^{\text{root}} \rVert$ & $-50$ \\
        DoF position & $\exp(-0.25\lVert \mathbf{\hat{p}} - \mathbf{p} \rVert_2)$ & $100$ \\  
        DoF velocity & $\exp(-0.25\lVert \mathbf{\hat{\dot{p}}} - \mathbf{\dot{p}} \rVert_2^2)$ & $10$ \\  
        Extend body position & $\exp(-0.5\lVert \mathbf{\hat{q}} - \mathbf{q} \rVert_2^2)$ & $250$ \\  
        Body position (\ac{mr}) & $\exp(-0.5\lVert \mathbf{q}_{\text{vr}} - \mathbf{\hat{q}}_{\text{vr}} \rVert_2^2)$ & $150$ \\  
        Body rotation & $\exp(-0.1\lVert \mathbf{\theta} \ominus \mathbf{\hat{\theta}} \rVert)$ & $400$ \\  
        Body velocity & $\exp(-10.0\lVert \mathbf{v} - \mathbf{\hat{v}} \rVert_2)$ & $80$ \\  
        Body angular velocity & $\exp(-0.01\lVert \boldsymbol{\omega} - \boldsymbol{\hat{\omega}} \rVert_2)$ & $8$ \\  
        Body hand rotation & $(\theta_{\text{hand}} - \hat{\theta}_{\text{hand}})^{2}$ & $500$ \\  
        AMP & \cref{sec:reward} & $500$ \\  
    \bottomrule 
    \end{tabular}
\end{table}

\begin{table}[ht!]
    \centering
    \small
    \caption{\textbf{Domain Randomization.} The details of the primary domain randomization used in our training process.}
    \label{tab:dr}
    \begin{tabular}{ c c }
        \hline
        Term & Value \\ 
        \hline
        Friction & $\mathcal{U}(0.6, 2.0)$ \\
        Base CoM offset & $\mathcal{U}(-0.04, 0.04) \text{m}$ \\
        Link mass & $\mathcal{U}(0.7, 1.25) \times \text{default} \ \text{kg}$ \\
        P Gain & $\mathcal{U}(0.85, 1.15) \times \text{default}$ \\
        D Gain & $\mathcal{U}(0.85, 1.15)  \times \text{default} $ \\
        Torque RFI & $0.05 \times \text{torque limit}\ \text{N}\cdot\text{m}$ \\
        Control delay & $\mathcal{U}(0.0, 20)\text{ms}$ \\ 
        Global step delay & $\mathcal{U}(0.0, 80)\text{ms}$ \\ 
        Rand born distance  & $\mathcal{U}(0.0, 2.0)\text{m}$ \\ 
        Rand heading degree  & $\mathcal{U}(-20.0, 20.0)\text{degree}$ \\
        Push robot & $\text{interval}=5s$, $v_{xy}=1.5 \text{m/s}$ \\ 
        Terrain type & flat, rough, low obstacles \cite{he2024omnih2o} \\ \hline
    \end{tabular}
\end{table}

\section{Implement Details}\label{sec:supp:impl}
\subsection{Data Augmentation}\label{sec:supp:data}
In our implementation, we filter out physically infeasible AMASS data and select motions with large upper- and lower-body workspaces as oracle motions. These motions are further modified by concatenating body parts or accelerating sequences.

\subsection{Model Architecture}\label{sec:supp:model}
The student policy is composed of $L = 3$ \ac{moe} layers, each containing $N = 4$ experts, where each expert is implemented as an \ac{mlp} with dimensions ($2048$, $512$, $512$, $256$).
The policy uses a history length of $H = 25$ frames and activates the top $k = 2$ experts based on the highest weights determined by the router. The AMP discriminator is a $3$-layer \ac{mlp} ($256$, $256$, $256$) that is updated online during training on \dataset. For comparison, the baseline model \model\textsuperscript{\(\dagger\)} uses a single \ac{mlp} with architecture ($2048$, $1024$, $512$, $512$).

\subsection{Policy Training}\label{sec:supp:training}
We train our policy in IsaacGym using a single A$800$ GPU. The teacher policy is trained for $1$M iterations with $8192$ parallel environments, while the student policy is trained for $600K$ iterations with $4096$ parallel environments. 
Training the teacher policy requires $\sim480$K simulation steps ($\sim20$K PPO steps) and $\sim24$ hours on a single A$800$ GPU. The student policy requires $\sim48$ hours on a single $3090$ Ti.

\section{Experiments}

\subsection{Evaluation Metrics}\label{sec:supp:metric}
We evaluated \model on motion tracking tasks from \dataset using five metrics: success rate $\mathbf{SR}$ (\%), mean per-keybody position error (MPKPE) $E_{\text{mpkpe}}$ (mm), root-relative mean per-keybody position error (R-MPKPE) $E_{\text{r-mpkpe}}$ (mm), average joint velocity error $E_{\text{vel}}$ (mm/s), and hand orientation tracking error $E_{\text{hand}}$. Success rate ($\mathbf{SR}$) represents the proportion of episodes where: (i) the robot maintains balance without falling, and (ii) the average per-keybody distance between the robot and reference motion remains below $1.5$m across the three controlled joints. We defined the hand orientation tracking error as $E_\text{hand}=1-\left<\hat{\textrm{q}}, \textrm{q}\right>^2$, where $\hat{\textrm{q}}$ and $\textrm{q}$ represent the reference and robot hand quaternions.

\subsection{Ablation Study}\label{sec:supp:ablation}
\begin{wraptable}{r}{0.48\linewidth}
    \centering
    \vspace{-9pt}
    \captionof{table}{Ablation study on history length and architecture components.}
    \scriptsize
    \begin{tabularx}{0.95\linewidth}{@{}l *{4}{>{\centering\arraybackslash}X}@{}}
        \toprule
        \textbf{Method} & \textbf{$E_{\text{mpkpe}}$}  & \textbf{$E_{\text{r-mpkpe}}$} & \textbf{$E_{\text{vel}}$} & \textbf{$E_{\text{hand-rot}}$} \\
        \midrule
        \rowcolor{gray!15}
        \multicolumn{5}{@{}l}{\textbf{(a) History Length Analysis}} \\
        \midrule
        History5  & $93.97$  & $\textbf{31.99}$  & $236.12$ & $3.80$ \\
        History50 & $135.60$ & $41.33$  & $286.66$ & $12.23$ \\
        History25(\model) & $\textbf{87.84}$ & $33.30$ & $\textbf{227.17}$ & $\textbf{3.61}$ \\
        \midrule
        \rowcolor{gray!15}
        \multicolumn{5}{@{}l}{\textbf{(b) Architecture Ablation}} \\
        \midrule
        \model ($L=1$)      & $134.06$  & $37.56$  & $270.14$ & $7.22$ \\
        \model ($N=8$)      & $89.21$  & $\textbf{30.90}$  & $251.10$ & $4.26$ \\
        \model & $\textbf{87.84}$ & $33.30$ & $\textbf{227.17}$ & $\textbf{3.61}$ \\
        \bottomrule
    \end{tabularx}
    \label{tab:ablation}
\end{wraptable}
We investigated the impact of key design choices, specifically history length and \ac{moe} parameters, through systematic ablation experiments reported in \cref{tab:ablation}. Our results indicate that a configuration using $25$ timesteps of history, three \ac{moe} layers, and four experts per layer yields optimal performance across most evaluation metrics. We observed that shorter history lengths and increased expert counts can produce marginally lower R-MPKPE values and larger global tracking errors, suggesting a trade-off between local and global motion fidelity.


\subsection{Qualitative Results Comparsion}\label{sec:supp:compare}

\begin{figure}[t!]
    \centering
    \includegraphics[width=\linewidth]{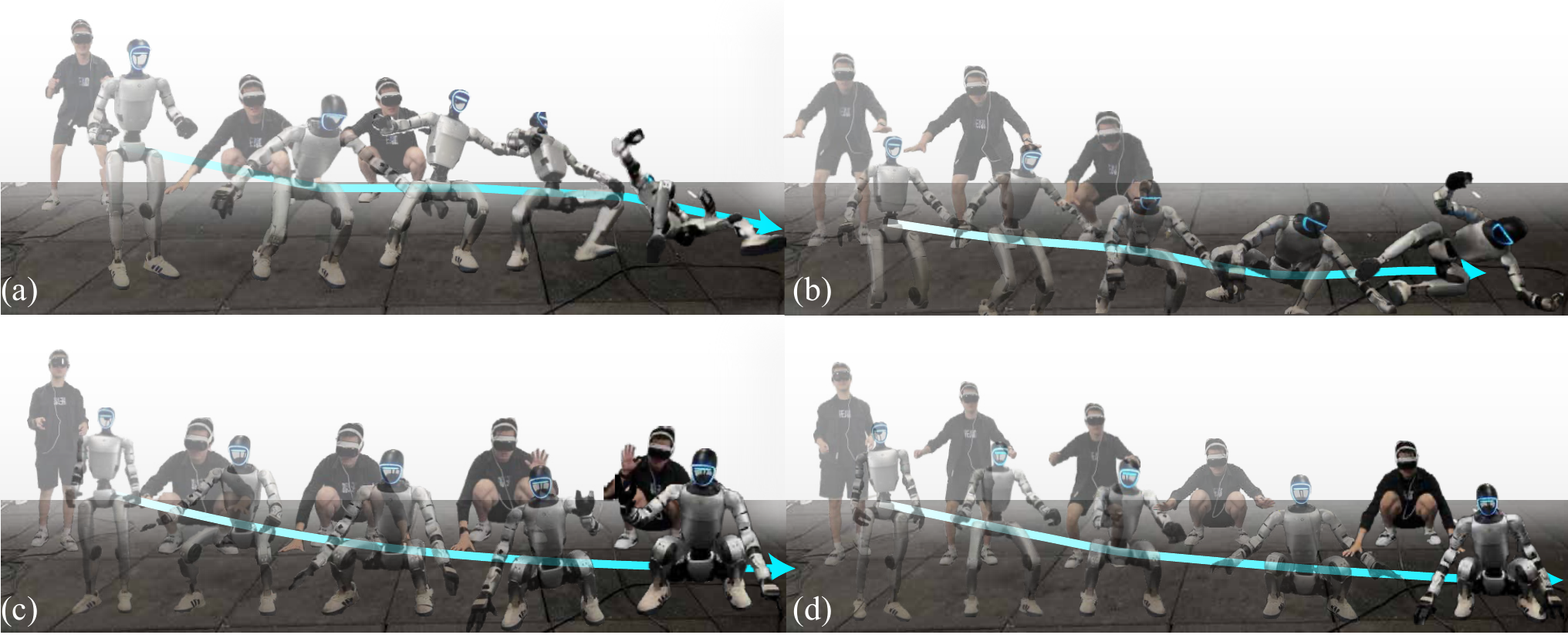}
    \caption{Qualitative Results of \model and \model\textsuperscript{*}. (a) and (b) show the ``crouch'' tracking results of \model\textsuperscript{*}, while (c) and (d) present the results of \model.}
    \label{fig:supp:quatitative}
\end{figure}

We analyze the qualitative results of \model and \model\textsuperscript{*} in \cref{fig:supp:quatitative}. Subfigures (a) and (b) show that \model\textsuperscript{*}, trained on OmniH2O \cite{he2024omnih2o}, fails to track motions like ``crouch'' or ``squat to pick up an object'' and falls down. In contrast, subfigures (c) and (d) present the results of \model, which tracks these motions accurately and robustly. Although \model\textsuperscript{*} is trained on a larger dataset (more than $8k$ motions, compared to \dataset's $345$ motions), it struggles with these tasks. Meanwhile, our model effectively tracks these motions and performs manipulation skills using only about $20\%$ of the data. Since the OmniH2O \cite{he2024omnih2o} dataset also includes motions like ``squat,'' this result suggests that a smaller dataset can still yield excellent tracking performance, as large-scale training data may cause the policy to overly generalize and compromise certain skills.

\paragraph{Expert Activation Analysis}
\begin{figure}
    \centering
    \includegraphics[width=\linewidth]{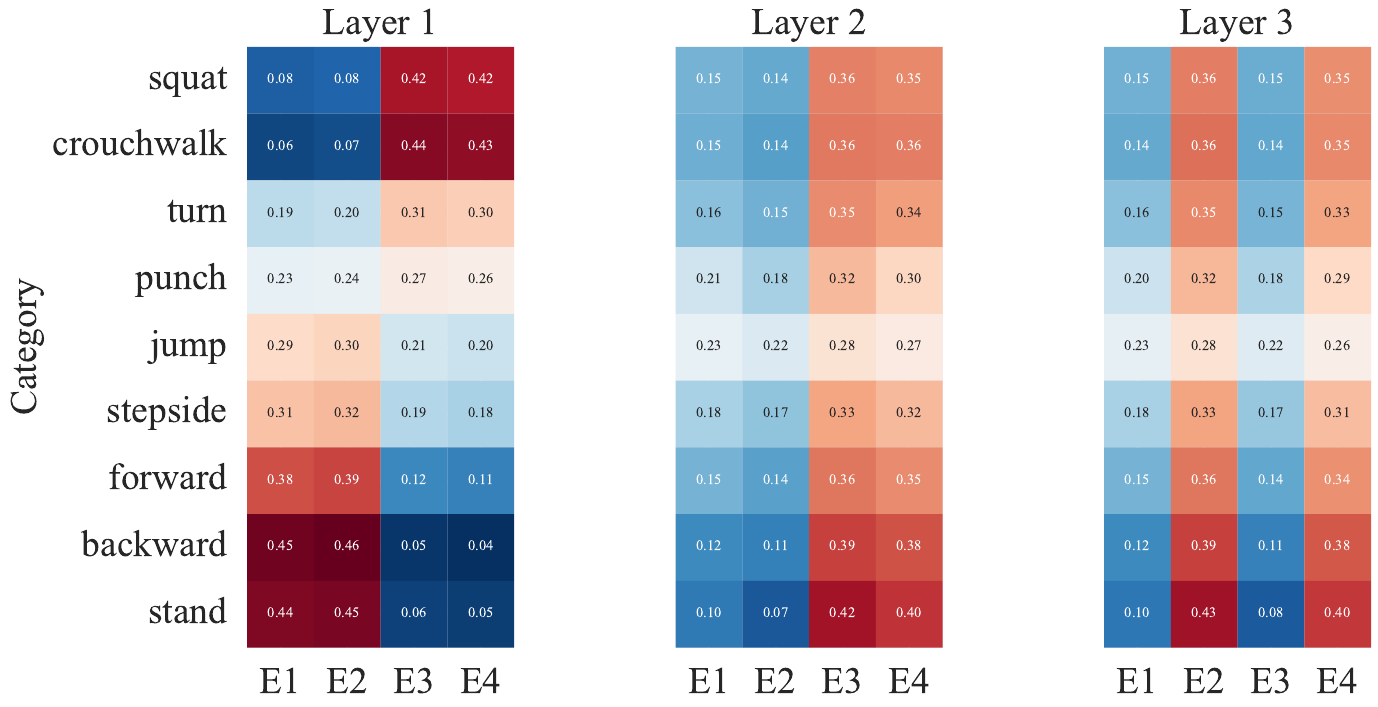}
    \caption{The activation status of each expert.}
    \label{fig:expert}
\end{figure}

To better understand the specialization within our mixture-of-experts architecture, we visualized expert activation weights across nine distinct motion types in \cref{fig:expert}. Results reveal clear specialization patterns where motions requiring similar skills activate specific experts. In the first layer, experts $1$ and $2$ are predominantly activated during standing motions, while experts $3$ and $4$ show stronger activation during squatting motions. Notably, all four experts in the first layer become activated during dynamic motions such as jumping and punching, suggesting collaborative processing of complex movements. Similar specialization patterns emerged in subsequent layers, albeit with reduced variance across different motion categories.

\subsection{The Choice of the Number of MoE Layers and Number of Experts}\label{sec:supp:moe}

\begin{figure}[t!]
    \centering
    \includegraphics[width=\linewidth]{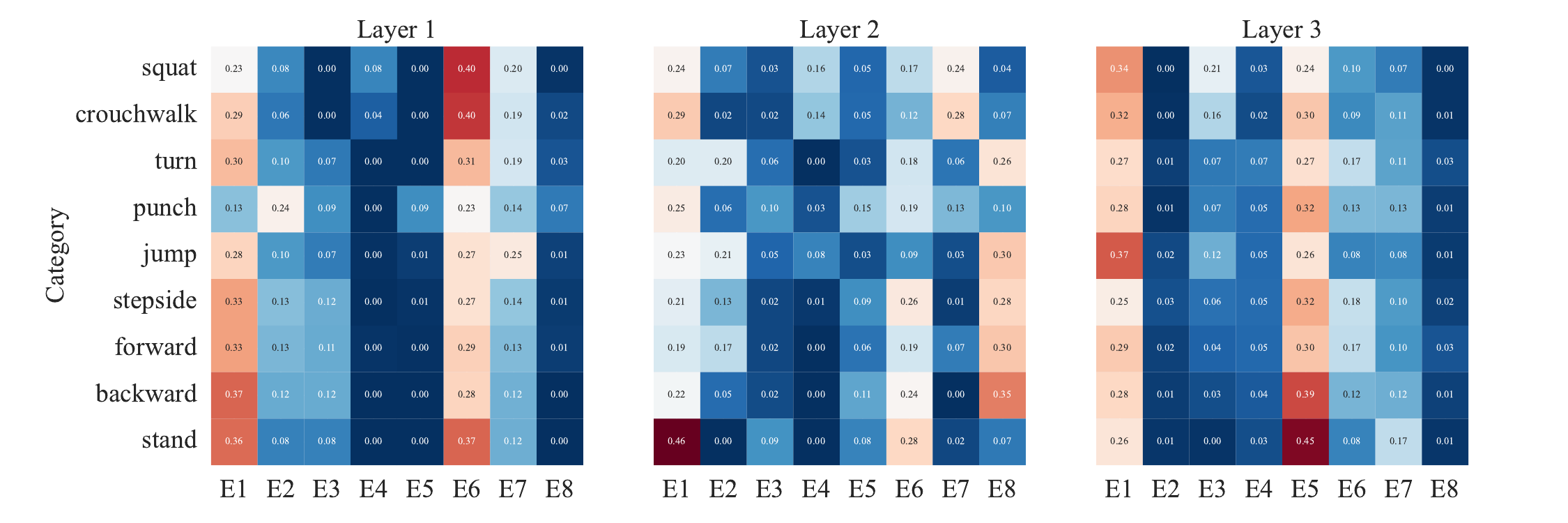}
    \caption{Experts activation when $N=8$}
    \label{fig:supp:expert_n}
\end{figure}

\begin{figure}[ht!]
    \centering
    \centering
    \includegraphics[width=\linewidth]{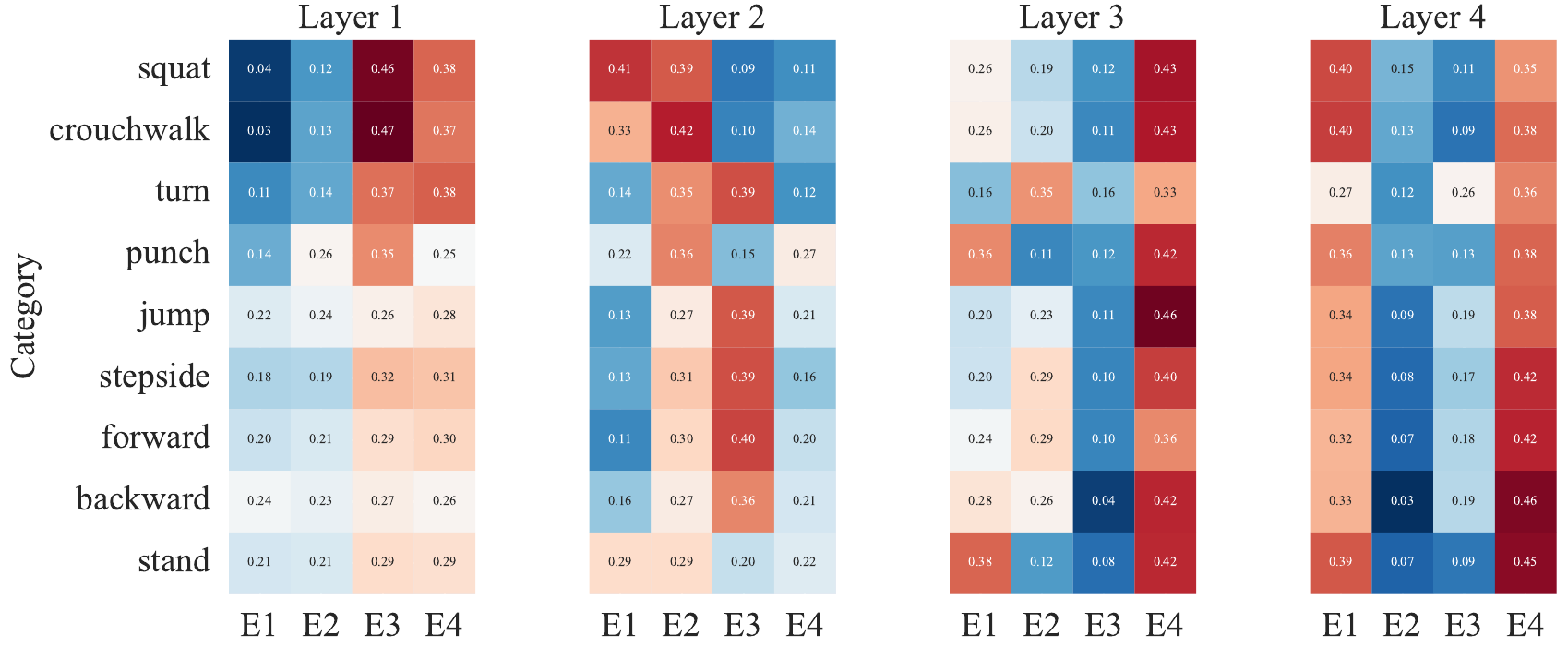}
    \caption{Experts activation when $L=4$}
    \label{fig:supp:expert_4}
\end{figure}

\begin{figure}[ht!]
    \centering
    \centering
    \includegraphics[width=\linewidth]{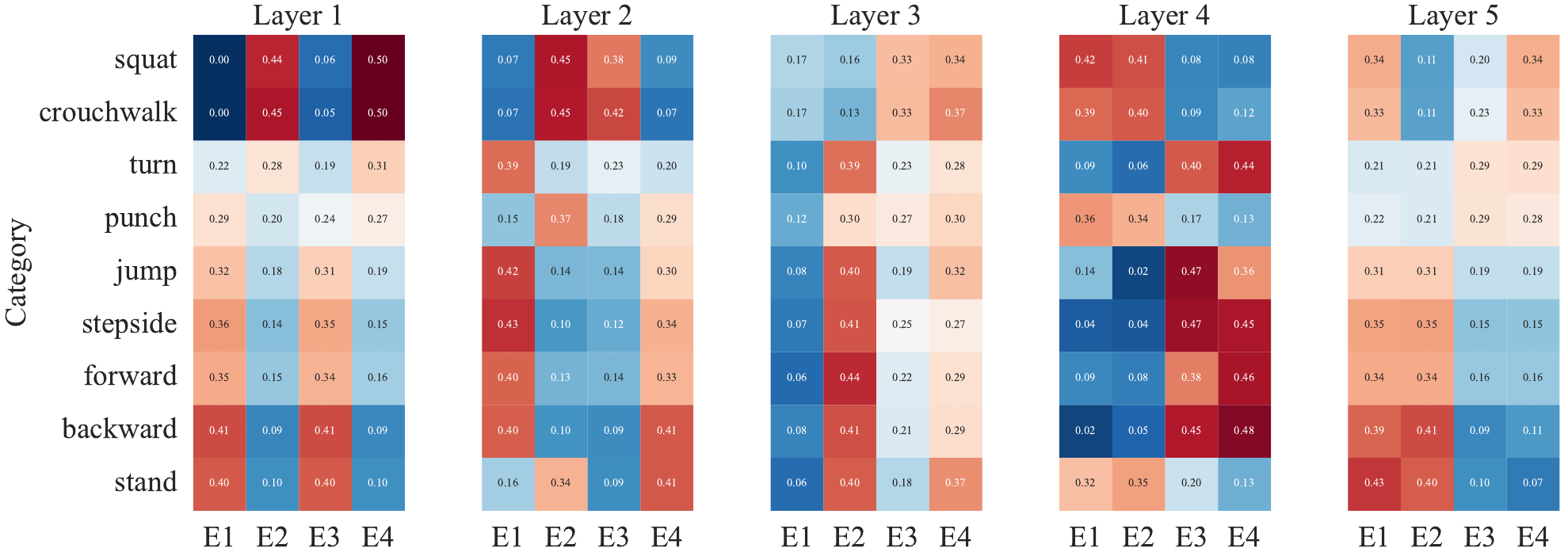}
    \caption{Experts activation when $L=5$}
    \label{fig:supp:expert_5}
\end{figure}

\begin{figure}
    \centering
    \includegraphics[width=0.4\linewidth]{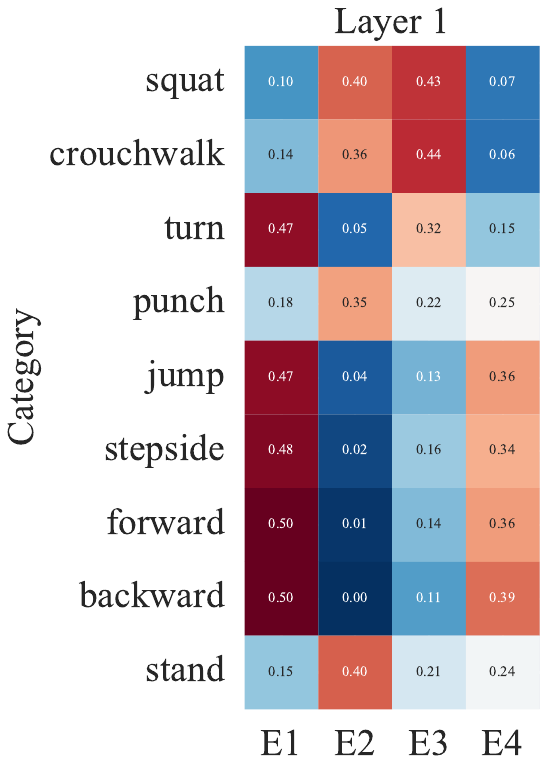}
    \caption{Experts activation when $L=1$}
    \label{fig:supp:expert_l}
\end{figure}

We visualize the activation patterns of experts in \cref{fig:expert,fig:supp:expert_n,fig:supp:expert_l,fig:supp:expert_4,fig:supp:expert_5}.
\cref{fig:supp:expert_n} shows that \ac{moe} layers with $N=8$ experts activate only half of the experts in each layer, revealing that $8$ experts are redundant for the current training data distribution, while $4$ experts are sufficient.
\cref{fig:supp:expert_l} demonstrates that \model\textsuperscript{*}($L=1$), which uses only one \ac{moe} layer, is still capable of activating different experts. However, as shown in \cref{tab:ablation}, its tracking performance is inferior to that of \model. This is primarily attributed to the model's parameters being too limited to effectively learn such diverse motions. Though $4$ \ac{moe} layers and $5$ \ac{moe} layers also has same activation patterns, like shown in \cref{fig:supp:expert_4,fig:supp:expert_5}, we choose $3$ \ac{moe} layers for a balance of training cost and policy performance. Therefore, we select the \ac{moe} policy with three \ac{moe} layers and four experts as our final model. 

\end{document}

%% file: config.tex
\usepackage{graphicx} \graphicspath{{figures/}}
\usepackage{times,latexsym}
\usepackage{amsmath}
\usepackage{amssymb}
\usepackage{acronym}
\usepackage{balance}
\usepackage{xspace}
\usepackage{setspace}
\usepackage[skip=3pt,font=small]{subcaption}
\usepackage[skip=3pt,font=small]{caption}
\usepackage{booktabs,tabularx,multirow,array,makecell}
\usepackage{overpic,wrapfig}
\usepackage[linesnumbered,ruled,vlined]{algorithm2e}
\usepackage{enumitem}
\usepackage{physics}
\usepackage[capitalise,nameinlink]{cleveref}
\usepackage{marvosym}

\def\model{\textbf{\textsc{Clone}}\xspace}
\def\dataset{\textbf{\textsc{Cloned}}\xspace}
\newcommand{\dna}{\includegraphics[height=2.8ex]{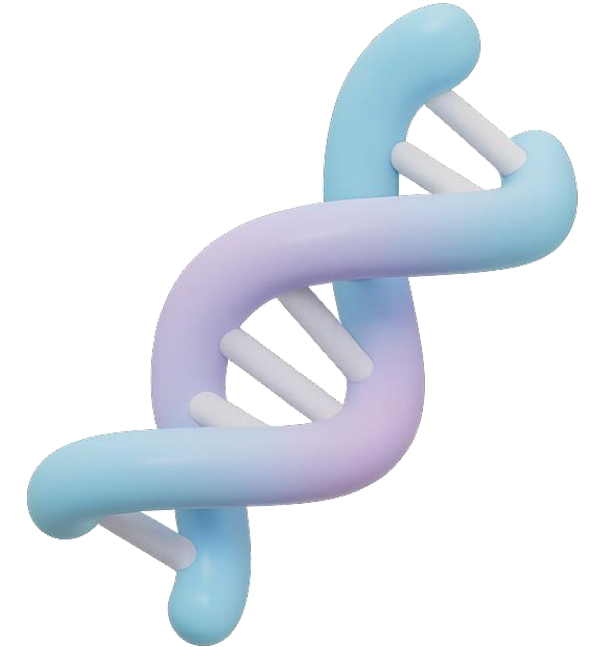}\hspace{2px}}

\makeatletter
\DeclareRobustCommand\onedot{\futurelet\@let@token\@onedot}
\def\@onedot{\ifx\@let@token.\else.\null\fi\xspace}
\def\eg{\emph{e.g}\onedot}

\def\vs{\emph{vs}\onedot}

\makeatother

\acrodef{mlp}[MLP]{Multi-Layer Perceptrons}
\acrodef{mdp}[MDP]{Markov Decision Process}
\acrodef{imu}[IMUs]{Inertial Measurement Units}
\acrodef{amp}[AMP]{Adversarial Motion Priors}
\acrodef{sde}[SDE]{Stochastic Differential Equation}
\acrodef{fk}[FK]{Forward Kinematics}
\acrodef{moe}[MoE]{Mixture-of-Experts}
\acrodef{slerp}[SLERP]{Spherical Linear Interpolation}
\acrodef{ppo}[PPO]{Proximal Policy Optimization}
\acrodef{cvae}[CVAE]{Conditional Variational Autoencoder}
\acrodef{mocap}[MoCap]{motion capture}
\acrodef{avp}[AVP]{Apple Vision Pro}
\acrodef{mr}[MR]{Mixed Reality}

\crefname{algorithm}{Alg.}{Algs.}
\Crefname{algocf}{Algorithm}{Algorithms}
\crefname{section}{Sec.}{Secs.}
\Crefname{section}{Section}{Sections}
\crefname{table}{Tab.}{Tabs.}
\Crefname{table}{Table}{Tables}
\crefname{figure}{Fig.}{Fig.}
\Crefname{figure}{Figure}{Figure}